%% file: main_acl.tex
\documentclass[11pt]{article}

\usepackage[final]{acl}

\usepackage{times}
\usepackage{latexsym}

\usepackage[T1]{fontenc}

\usepackage[utf8]{inputenc}

\usepackage{microtype}

\usepackage{inconsolata}

\usepackage{graphicx}

\title{CrochetBench: Can Vision-Language Models Move from Describing to Doing in Crochet Domain?}
\author{Peiyu Li\thanks{These authors contributed equally to this work.}, Xiaobao Huang\protect\footnotemark[1], Ting Hua, Nitesh V. Chawla\\
        University of Notre Dame\\
        \texttt{\{pli9, xhuang2\}@nd.edu} }

\input{math_commands.tex}

\usepackage{booktabs,tabularx,makecell,multirow,xcolor,colortbl}
\usepackage{hyperref}
\usepackage{url}
\usepackage{multirow, multicol}
\usepackage{array}
\usepackage{makecell} 
\usepackage{booktabs}
\usepackage{tikz}
\usepackage{arydshln} 
\usepackage[most]{tcolorbox}
\tcbuselibrary{breakable}
\usepackage{graphicx}
\usepackage{wrapfig}
\usepackage{subcaption}
\usetikzlibrary{trees}
\usepackage{float}

\usepackage{makecell}

\usepackage{graphicx}
\usepackage{subcaption}
\usepackage[most]{tcolorbox} 
\usepackage{ragged2e}        

\usepackage{colortbl}
\definecolor{qwenlight}{RGB}{235,245,255}
\definecolor{qwendark}{RGB}{210,230,255}

\newcommand{\caseblock}[3]{%
  \begin{subfigure}{0.16\textwidth}
    \centering
    \includegraphics[width=\linewidth]{#2}
    \vspace{1pt}
    
    \scriptsize
    \begin{tcolorbox}[
  colback=gray!5,
  colframe=gray!40,
  boxrule=0.3pt,
  left=2pt,right=2pt,top=2pt,bottom=2pt,
  height=4.1cm,
  valign=top
]
      \RaggedRight\ttfamily
      #3
    \end{tcolorbox}
    
    \caption{#1}
  \end{subfigure}%
}

\begin{document}

\maketitle

\begin{abstract}

While multimodal large language models can describe visual content, their ability to generate executable procedures remains underexplored. CrochetBench presented in this paper evaluates this shift from \emph{describing} to \emph{doing} through fine-grained procedural reasoning in crochet: models must recognize stitches, select structurally appropriate instructions, and generate compilable procedures. We adopt the \emph{CrochetPARADE DSL} as our intermediate representation, enabling structural validation and functional evaluation via execution. The benchmark covers tasks including stitch classification, instruction grounding, and both natural language and image-to-DSL translation. Across all tasks, performance sharply decreases as the evaluation shifts from surface-level similarity to executable correctness, revealing limitations in long-range symbolic reasoning and 3D-aware procedural synthesis. Our proposed CrochetBench offers a new lens for assessing procedural competence in multimodal models and highlights the gap between surface-level understanding and executable precision in real-world creative domains. Code is available at \url{https://github.com/Peiyu-Georgia-Li/crochetBench}. 
\end{abstract}

\input{section/intro}

\input{section/data_description}

\input{section/task_formulation}

\input{section/experiment}

\input{section/related_work}

\section{Conclusion}

 
 CrochetBench provides a structured benchmark for assessing whether multimodal LLMs can move from recognizing visual content to executing the step-by-step procedures required to produce a crochet pattern. Across all four tasks, models demonstrate a consistent gap: they can identify stitches and retrieve plausible instructions, but they fail to generate structurally valid procedures or produce executable programs.
 Qualitative analysis reveals that fluent outputs often misrepresent global structure, and scaling further exacerbates execution failures due to uncontrolled invention of stitch types. Supervised finetuning improves text-based metrics but does not yield corresponding gains in executable correctness. These results highlight the need for learning approaches that explicitly target structured and long-horizon procedural reasoning.
\clearpage
\newpage

\section*{Limitations}
While CrochetBench enables executable evaluation of visual procedural reasoning, it has several limitations. First, the benchmark focuses on a single creative domain, crochet, which although representative of structured symbolic procedures, may not capture all forms of procedural reasoning found in domains such as robotics, mechanical assembly, or scientific experimentation. As a result, generalization to other domains should be interpreted with caution.

Second, executable correctness is assessed through the CrochetPARADE domain specific language, which necessarily abstracts certain aspects of real world crochet practice. Multiple distinct procedures can produce visually similar artifacts, and some valid human written patterns may be penalized due to constraints of the language or limitations of the rendering process. Although we mitigate this issue using functional validation and image based similarity metrics, these signals remain imperfect proxies for true semantic equivalence.

Third, we do not explicitly control for potential overlap with model pretraining corpora. While such exposure is a realistic condition for web-scale benchmarks, we cannot fully rule out its influence on model performance.

Finally, while we evaluate a diverse set of state-of-the-art vision language models, our finetuning experiments are limited to supervised learning on a single architecture. Exploring alternative training paradigms, such as program guided learning or learning with execution feedback, remains an important direction for future work.

\section*{Ethical Considerations}
We take licensing considerations seriously and obtained explicit confirmation from the copyright holder (Yarnspirations) that the patterns may be used for research and personal purposes, while reproduction for resale is not permitted. Accordingly, the dataset is designed to avoid redistribution of original content. We release only structured JSON annotations, reference URLs to the original sources, and the parsing and annotation scripts. This approach enables reproducible research while respecting intellectual property. The benchmark is intended strictly for non-commercial academic use.



\bibliography{iclr2026_conference}

\clearpage
\appendix
\input{appendix/crochet_primer}
\input{appendix/crochetBench_analysis}

\input{appendix/prompt}

\input{appendix/bertscore}

\input{appendix/crochetparade}

\input{appendix/error_type}

\input{appendix/case_study_example}
\end{document}

%% file: math_commands.tex
\usepackage{amsmath,amsfonts,bm}

\def\eqref#1{equation~\ref{#1}}

\def\1{\bm{1}}

\DeclareMathAlphabet{\mathsfit}{\encodingdefault}{\sfdefault}{m}{sl}
\SetMathAlphabet{\mathsfit}{bold}{\encodingdefault}{\sfdefault}{bx}{n}



%% file: section/intro.tex
\begin{figure*}[t]
    \centering
    \includegraphics[width=\linewidth]{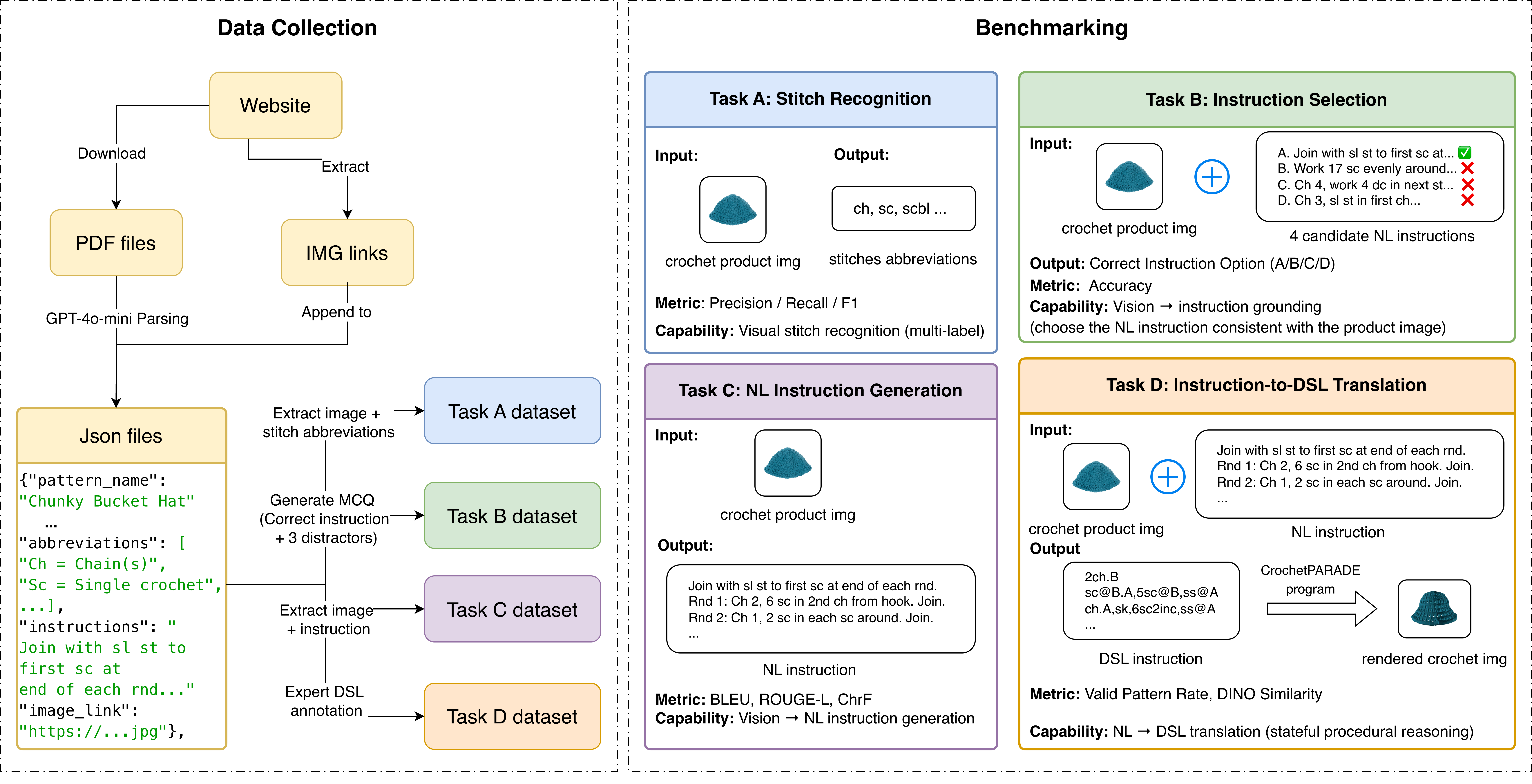}
    \caption{\textbf{End-to-end data construction and benchmarking workflow for CrochetBench.} The left panel illustrates the \emph{data collection pipeline}: we  download PDF files and image links from crochet pattern websites, and parse them using GPT-4o-mini to produce structured JSON files. From each JSON record, we derive four supervised datasets: (A) stitch-level labels, (B) multiple-choice instruction selection items, (C) natural-language instruction generation pairs, and (D) expert-annotated DSL programs for procedural synthesis. The right panel summarizes the four benchmarking tasks: \textbf{Task A} evaluates multi-label visual stitch recognition; \textbf{Task B} measures vision-to-instruction grounding via MCQ selection; \textbf{Task C} assesses vision-conditioned natural-language instruction generation; and \textbf{Task D} tests stateful procedural reasoning via NL-to-DSL (Natural Language to Domain-Specific Language) translation with execution-based metrics.}
    \label{fig:crochetbench_pipeline}
\end{figure*}

\section{Introduction}
Multimodal large language models can describe what they see, but can they
explain how to make it?   
Given an image of a piece of furniture or a crocheted blanket, current models can
fluently describe its appearance, yet cannot generate the step-by-step instructions
needed to assemble or recreate it. Closing this gap, from
perception to executable procedure, is critical for applications in robotics,
CAD design, and procedural content generation. However, evaluating procedural
competence is difficult: most real-world procedures require physical execution
to verify correctness. 
One domain offers a rare exception: crochet instruction synthesis. 
These crochet instructions follow a formal symbolic
grammar that can be compiled and validated automatically, enabling scalable
evaluation of procedural competence without requiring costly physical execution.

Existing multimodal benchmarks largely focus on descriptive tasks. Image 
captioning benchmarks such as COCO \citep{lin2015microsoftcococommonobjects}  and TextCaps \citep{sidorov2020textcapsdatasetimagecaptioning} evaluate whether models can generate natural language descriptions of visual scenes. 
Text-to-code benchmarks such as HumanEval \citep{chen2021evaluatinglargelanguagemodels} evaluate program synthesis but lack visual grounding, while image-to-code 
benchmarks such as pix2code \citep{beltramelli2017pix2code} test layout generation rather 
than stateful procedural reasoning. 
Even procedural datasets such as
Recipe1M+ \citep{marin2019recipe1mdatasetlearningcrossmodal} primarily assess cross-modal alignment rather
than executable correctness, since validating a recipe requires cooking it, a slow
and resource-intensive process \citep{chefFusion}. 
No existing benchmark combines visual understanding, multi-step procedural
decomposition, and automated executable validation.

To address this gap, we turn to crochet: an ideal testbed for visual procedural
reasoning. Each crochet pattern presents a vision-to-procedure challenge. 
Given an image of a finished artifact, models must infer the sequential construction process: which stitches were used, in what order, and with what counts. This requires
(1) fine-grained visual recognition of stitch types from texture and topology,
(2) procedural decomposition into ordered operations with symbolic grounding,
(3) stateful reasoning to track counts across steps, and
(4) constraint satisfaction to ensure topological consistency.
What makes crochet unique is that, crochet instructions can be formalized into domain-specific languages and validated through compilation, which enables the automated executable evaluation that other procedural domains lack.

In this paper, we introduce \textbf{CrochetBench}, the first benchmark for executable visual
procedural reasoning in creative domains. 
Our CrochetBench comprises 6,085 crochet patterns spanning 55 categories with over 98\% image coverage, ranging from
beginner to expert complexity. 
We design four tasks of increasing difficulty:
\textbf{Task A (Stitch Recognition)} evaluates multi-label visual classification;
\textbf{Task B (Instruction Selection)} evaluates vision-to-language grounding via
multiple-choice matching;
\textbf{Task C (Instruction Generation)} evaluates free-form instruction synthesis
from images; and
\textbf{Task D (DSL Translation)} evaluates translation into executable
CrochetPARADE programs~\citep{crochetparade}.
Together, these tasks shift evaluation from surface-level fluency to procedural
correctness, whether outputs compile and preserve structural constraints.

We evaluate nine state-of-the-art vision-language models on CrochetBench, including GPT-4o \citep{hurst2024gpt},
Gemini 2.5 Flash-Lite \citep{comanici2025gemini}, Claude Sonnet 4 \citep{anthropic2025claude_sonnet4}, and open-source alternatives spanning
3B to 72B parameters.

Our evaluation reveals that current models exhibit a consistent gap between
perception and procedural synthesis. 
Models achieve moderate performance on stitch recognition and instruction selection,
but struggle to generate multi-step instructions or executable DSL programs. 
A case study (Figure~\ref{fig:case_study_main})
illustrates this pattern: even when models produce fluent instructions that
correctly identify local details such as colors and motifs, the resulting
procedures misconstruct global geometry and fail under execution. 
Also, neither model scaling nor supervised finetuning closes this gap. Larger models often achieve lower valid pattern rates during execution due to a tendency to invent novel stitch types not defined in the DSL, while finetuning improves text-based metrics without improving executable correctness.

\textbf{Our contributions} are threefold:
(1) We release CrochetBench, the first executable benchmark for procedural textile
crafts, comprising 6,085 patterns integrated with the CrochetPARADE DSL for
automated validation beyond text similarity.
(2) We design four tasks of increasing difficulty, including stitch recognition, instruction
selection, instruction generation, and DSL translation, enabling fine-grained
diagnosis of procedural competence.
(3) We evaluate nine state-of-the-art vision–language models on CrochetBench and reveal a consistent
gap between recognition and executable synthesis that persists across model scales
and finetuning.

%% file: section/data_description.tex
\section{Dataset Description}

\textbf{CrochetBench} is a large-scale, structured benchmark comprising 6,085 crochet patterns across 55 project categories. As shown in Figure~\ref{fig:crochetbench_pipeline} (left), the dataset is constructed from publicly available patterns on the Yarnspirations website,\footnote{\url{https://www.yarnspirations.com/collections/patterns}} a widely used repository in the fiber-arts community. The raw patterns, originally distributed as PDFs, are parsed and normalized via a GPT-4o-mini–based pipeline that extracts and standardizes key fields such as  metadata, materials, measurements, gauge, abbreviations, and step-by-step instructions. Each pattern is paired with a high-resolution finished-product image (typically 2000$\times$2000 pixels, sRGB) and represented as a structured JSON object under a unified schema. To assess the reliability of the parsing pipeline, we conducted a focused validation using both quantitative overlap analysis and manual inspection. Across 494 randomly sampled patterns spanning 55 categories, we observe 97.1\% word-level overlap between parsed JSON and original PDFs, and 95–100\% overlap specifically within procedural instruction sections, confirming high extraction fidelity.

CrochetBench exhibits substantial diversity in both visual appearance and procedural complexity. Patterns vary widely in instruction length, stitch vocabulary, and structural depth, supporting evaluation across different levels of difficulty and long-horizon reasoning. Detailed statistics, skill-level distributions, and a JSON example are provided in the Appendix~\ref{app:data-statistics}.

%% file: section/task_formulation.tex
\section{Tasks}
\label{sec:tasks}
A central goal of CrochetBench is to evaluate whether multimodal LLMs can move beyond surface-level visual description and produce \emph{procedurally correct} crochet instructions. As shown in Table~\ref{tab:crochetbench_tasks}, CrochetBench is organized as a progression of four tasks that isolate the core cognitive abilities required for real-world crochet reasoning. 

Tasks~A and~B focus on \textbf{perception and comprehension}, representing the minimum prerequisites for procedural understanding. Stitch recognition and instruction selection evaluate whether models can identify stitch types from images and correctly associate visual patterns with corresponding natural-language instructions. However, accomplishing these two tasks does not guarantee the ability to synthesize a valid crochet procedure. Tasks~C and~D therefore target \textbf{procedural generation and formalization}, requiring models to produce coherent, stepwise natural-language instructions or executable CrochetPARADE programs. These tasks demand the integration of visual grounding, temporal consistency, symbolic manipulation, and domain-specific constraints. 
The following subsections describe each task in detail.



\begin{table*}[t]
\centering
\small
\setlength{\tabcolsep}{5pt}
\renewcommand{\arraystretch}{1.15}

\begin{tabularx}{\textwidth}{@{}c l l X c l@{}}
\toprule
\textbf{ID} & \textbf{Ability Tested} & \textbf{Task} & \textbf{Evaluation Metrics} & \textbf{Dataset Size} & \textbf{Eval Set} \\
\midrule
A & Recognition & Stitch Recognition & F1, Precision, Recall & 6,009 & CrochetBench-A \\
B & Comprehension & Instruction Selection & Accuracy & 6,003 & CrochetBench-B \\
C & Generation & Instruction Generation & BLEU, ROUGE, ChrF & 6,009 & CrochetBench-C \\
\addlinespace[2pt]
\multirow{2}{*}{D} & \multirow{2}{*}{Formalization}
& Instr.-to-DSL (Step) & Valid Pattern Rate & 119 & CrochetBench-D\textsubscript{step} \\
& & Instr.-to-DSL (Project) & Valid Pattern Rate, Dino Similarity & 6,009 & CrochetBench-D\textsubscript{proj} \\
\bottomrule
\end{tabularx}
\caption{CrochetBench task overview and evaluation metrics.}
\label{tab:crochetbench_tasks}
\end{table*}

\subsection{Task A: Stitch Recognition}
Task~A evaluates a model's ability to identify crochet stitch types from an image of a finished product. We construct \textbf{CrochetBench-A}, a subset of 6,009 examples from the full benchmark, where each product image is paired with ground-truth stitch annotations. These labels are derived from the official pattern instructions and normalized into a standardized set of stitch abbreviations (e.g., \texttt{sc}, \texttt{hdc}, \texttt{dc}) to ensure consistency across patterns. Unlike standard image classification, this is a \emph{multi-label prediction problem}: multiple stitches may co-occur within the same image, often with subtle visual differences in texture and geometry. This task therefore probes fine-grained visual grounding of structured crochet semantics.

\paragraph{Evaluation.} 
For each example, we compute overlap between the predicted and reference stitch sets. True Positives (TP) are stitches correctly predicted; False Positives (FP) are stitches predicted but not in the reference; and False Negatives (FN) are stitches in the reference but missed by the model. From these counts, we compute precision (fraction of correct predictions among all predictions), recall (fraction of ground-truth stitches recovered), and F1 score (harmonic mean) \cite{powers2020evaluation}. Metrics are averaged across examples to provide overall performance. This formulation rewards models that recover all present stitches while avoiding spurious predictions.

Accurate stitch recognition is foundational for the benchmark, as later tasks (e.g., instruction selection and instruction generation) depend on robust detection of stitch primitives.

\subsection{Task B: Instruction Selection}

Task~B evaluates whether a model can correctly associate an image of a finished crochet product image with its corresponding natural-language instruction. We construct \textbf{CrochetBench-B}, a subset of 6{,}003 examples, where each instance contains one ground-truth instruction and three distractor instructions sampled from the same project category (e.g., hats, rugs). Because distractors originate from the same category, they share similar visual and lexical structure, thereby increasing task difficulty and preventing solutions based on superficial lexical overlap. The answer distribution across options is approximately uniform (A: 24.9\%, B: 25.7\%, C: 23.7\%, D: 25.7\%), ensuring no positional bias.

\paragraph{Evaluation.}
To support scalable and reproducible benchmarking, we formulate the task as a four-way multiple-choice question (MCQ). The model must select one option (A–D), with exactly one correct answer. Predictions are extracted using a deterministic regex-based parser that identifies explicit letter-based responses (e.g., ``A'', ``Option B'', ``The answer is D''). Accuracy is used as the evaluation metric.

This task provides a controlled measure of visual grounding and semantic alignment between images and procedural text, without requiring free-form generation. By forcing discrimination among near-neighbor instructions, Task~B probes whether models can leverage fine-grained visual cues and domain-specific stitch semantics, which are essential precursors to reliable procedural instruction generation.

\subsection{Task C: Instruction Generation}

Task~C evaluates a model’s ability to generate natural-language crochet instructions from an image of a finished item. We construct \textbf{CrochetBench-C}, a subset of 6{,}009 examples in which each image is paired with the corresponding ground-truth textual instruction. In contrast to captioning or stylistic description, this task requires generating a sequence of domain-specific commands (e.g., ``Rnd 1: ch 4, 6 sc in ring''), each of which encodes precise stitch operations, counts, and ordering. Because real crochet patterns may include tens of steps, hierarchical structure (rounds, rows, substeps), and long-range dependencies, this task assesses whether models can infer the underlying procedural logic implied by the final visual product. The generated text must maintain consistent stitch semantics, preserve temporal ordering, and follow established formatting conventions used by human crafters.

\paragraph{Evaluation.}
We evaluate generation quality using BLEU, ROUGE-L, and ChrF \citep{papineni2002bleu,lin2004rouge,popovic2015chrf}, which together capture complementary aspects of textual fidelity in procedural instructions. BLEU measures overlap of word-level n-grams and thus reflects local lexical accuracy in stitch tokens and command sequences. ROUGE-L evaluates the longest common subsequence between the generated and reference patterns, capturing larger-scale ordering and structural alignment across multi-step procedures. ChrF operates on character-level n-grams, which makes it effective for crochet patterns where stitch abbreviations (e.g., \texttt{sc}, \texttt{sc2tog}) often differ by only a few characters. Word-based metrics treat such tokens as entirely distinct, whereas character-level comparisons can capture partial matches and small but semantically important variations. 

We do not adopt BERTScore \citep{zhang2020bertscore} as a primary metric. Although it captures semantic similarity via contextual embeddings, scores cluster within a narrow range (0.80–0.83 for strong models; see Appendix~\ref{app:bert_score}), limiting discriminative power. In the crochet setting, instructions are highly templated and lexically repetitive, so embedding similarity mainly reflects topical coherence (“this looks like a crochet pattern”) while remaining insensitive to structurally critical token-level differences, such as stitch counts, repeat multipliers, or round dependencies. 

However, textual overlap metrics alone cannot reveal whether the generated instructions form a coherent or executable procedure. A model may generate instructions that appear fluent and pattern-like while still violating fundamental structural constraints, including inconsistent stitch counts, infeasible transitions, or unbalanced repeat constructions. To directly assess structural correctness and program-level understanding, we introduce Task~D, which requires models to formalize correct natural-language instructions into a machine-checkable DSL representation.

\subsection{Task D: Instruction-to-DSL Translation}
Tasks~A–C evaluate perception, retrieval, and natural-language generation, but they do not test whether a model can represent crochet procedures in a structured, machine-interpretable form. Crochet patterns are inherently programmatic: they contain loops, repeats, and counting logic that natural language expresses only implicitly, and that text-based metrics cannot reliably validate. Task D isolates this structural dimension by requiring models to translate correct natural-language instructions into an executable DSL, thereby revealing whether models grasp the underlying program-like structure of crochet. This capability is essential for true procedural reasoning, and we instantiate it using the \textsc{CrochetPARADE} DSL. 

We construct two variants of Task~D: \textbf{CrochetBench-D\textsubscript{step}} (119 items) for step-level formalization and \textbf{CrochetBench-D\textsubscript{proj}} (6,009 items) for project-level program synthesis.

\paragraph{Step-Level Translation} 

The step-level task evaluates incremental translation from natural-language crochet instructions to DSL, where each instruction updates a shared stitch state. In this setting, the model is given a prefix of correct NL--DSL pairs representing the pattern translated so far and must generate the corresponding DSL line for the next instruction. Because crochet patterns are stateful, correct translation requires maintaining consistency with previous steps, including stitch counts, repeat structures, turning logic, and round-to-round topology.

This formulation tests whether models can map local textual cues into the structured symbolic operations of CrochetPARADE while preserving long-range procedural context. To capture variation in pattern progression, \textbf{CrochetBench-D\textsubscript{step}} groups examples by prefix length: 52 early-step examples use prefixes covering steps~1--2, 34 mid-step examples use prefixes covering steps~1--4, and 33 late-step examples use prefixes covering steps~1--6. Each example requires translating the next natural-language instruction given the corresponding NL--DSL history. This dataset is small, as each example requires fine-grained, line-by-line annotation in the CrochetPARADE DSL, with careful verification to ensure both syntactic validity and procedural correctness. This process demands substantial human effort and domain expertise, making large-scale annotation impractical while ensuring high-quality evaluation signals.

\paragraph{Project-Level Translation} 
In the project-level setting, the model is provided with the complete crochet instruction in natural language together with the corresponding product image, and must generate an entire CrochetPARADE program. This variant is globally self-contained but considerably more challenging than the step-level task: models must track stitch states over long horizons, resolve ambiguities in natural language, and produce code that is both syntactically valid and semantically aligned with the final design. This setting reflects how crochet instructions are used in practice, where each step depends on the correctness of all preceding steps. Image grounding is especially helpful for interpreting repeated motifs, symmetry, shaping, and termination conditions that may be under-specified in text alone.

\paragraph{Evaluation.}
Because crochet patterns are inherently free-form (where multiple distinct programs can yield the same final product and a single natural-language instruction may admit several semantically equivalent DSL realizations), there is no canonical gold program for Task~D. Exact string matching would therefore misjudge many correct solutions. Instead, CrochetBench evaluates correctness through \emph{functional executability} using the CrochetPARADE validator, which checks whether a predicted DSL program is syntactically valid, structurally consistent, and fully executable.

We use two complementary evaluation settings. For step-level translation, we report the \textbf{Valid Pattern Rate}, defined as the proportion of generated DSL steps that successfully compile. For project-level translation, we compute the \textbf{Valid Pattern Rate} for full programs and, for those that compile, render the executable portion into a crochet-like image and compute its \textbf{DINO Similarity} \citep{oquab2023dinov2} to the ground-truth product image, providing a coarse measure of semantic fidelity beyond syntax. To diagnose failure modes, we further identify the \textbf{first point of failure} for each invalid prediction and categorize it using our fine-grained error taxonomy (Appendix~\ref{app:dsl_error}), enabling us to distinguish local symbolic errors from broader state-tracking failures or misinterpretations of the natural-language instruction.

%% file: section/experiment.tex
\begin{table*}[ht]
\scriptsize 
\centering

\setlength{\tabcolsep}{5pt}
\begin{tabular}{llcccccccc}
\toprule
& \textbf{Model} & \textbf{Size} &
\multicolumn{3}{c}{\textbf{Stitch Recognition (\%)}} &
\textbf{Instr. Sel. (\%)} &
\multicolumn{3}{c}{\textbf{Instr. Gen. (\%)}} \\
\cmidrule(lr){4-6} \cmidrule(lr){7-7} \cmidrule(lr){8-10}
&  &  & \textbf{Prec} & \textbf{Rec} & \textbf{F1} & \textbf{Acc} & \textbf{BLEU} & \textbf{ROUGE-L} & \textbf{ChrF} \\
\midrule

\multirow{6}{*}{\shortstack{Open \\ Source}}
& BLIP-2 Flan-T5 XL & 3B
& 29.53 & 23.03 & 22.50
& 25.62
& 0.22 & 9.30 & 9.45 \\

& Google Gemma 3 & 4B
& 20.54 & 10.21 & 12.65
& 24.94
& 0.11 & 3.67 & 5.59 \\

& Google Gemma 3 & 27B
& 17.19 & 18.14 & 16.05
& 24.94
& 0.40 & 5.17 & 6.55 \\

& DeepSeek-VL & 7B
& 54.47 & \textbf{74.76} & \underline{60.60}
& 28.92
& 1.30 & 19.39 & 18.14 \\

 & \cellcolor{qwenlight}Qwen2-VL & \cellcolor{qwenlight}7B
& \cellcolor{qwenlight}54.14 & \cellcolor{qwenlight}\underline{69.74} & \cellcolor{qwenlight}58.16
& \cellcolor{qwenlight}41.96
& \cellcolor{qwenlight}1.67 & \cellcolor{qwenlight}21.10 & \cellcolor{qwenlight}15.99 \\

& Qwen2-VL & 72B
& 71.86 & 42.68 & 50.19
&  \textbf{68.85}
& 2.25 & 21.43 & 19.82 \\
\midrule

\multirow{3}{*}{\shortstack{Closed \\ Source}}
& GPT-4o & --
& 62.14 & 59.39 & 58.01
& \underline{58.11}
& 3.38 & 23.57 & \underline{24.07} \\

& Gemini 2.5 Flash-Lite & --
& \underline{74.49} & 49.77 & 56.83
& 55.63
& \underline{4.93} & \textbf{25.92} & \textbf{30.50} \\

& Claude Sonnet 4 & --
& \textbf{78.61} & 53.12 & \textbf{60.94}
& 57.39
& 3.35 & \underline{25.12} & 23.16 \\

\midrule

Finetuned
& \cellcolor{qwenlight}Qwen2-VL (finetuned)
& \cellcolor{qwenlight}7B
& \cellcolor{qwenlight}- 
& \cellcolor{qwenlight}- 
& \cellcolor{qwenlight}- 
& \cellcolor{qwenlight}- 
& \cellcolor{qwenlight}\makecell[c]{\textbf{5.64}\\{\scriptsize$\textcolor{blue}{(\uparrow +3.97)}$}}
& \cellcolor{qwenlight}\makecell[c]{25.10\\{\scriptsize$\textcolor{blue}{(\uparrow +4.00)}$}}
& \cellcolor{qwenlight}\makecell[c]{22.39\\{\scriptsize$\textcolor{blue}{(\uparrow +6.40)}$}} \\

\bottomrule
\end{tabular}
\caption{
Combined evaluation results across all three CrochetBench tasks:
\textit{Stitch Recognition}, \textit{Instruction Selection}, and \textit{Instruction Generation}.
Best values are \textbf{bold}; second-best are \underline{underlined}. For the finetuned Qwen2-VL 7B model, values in parentheses indicate absolute improvements over the pretrained Qwen2-VL 7B baseline.
}
\label{tab:combined_all}
\end{table*}

\begin{figure*}[t]
  \centering
  \caseblock{Ground Truth}{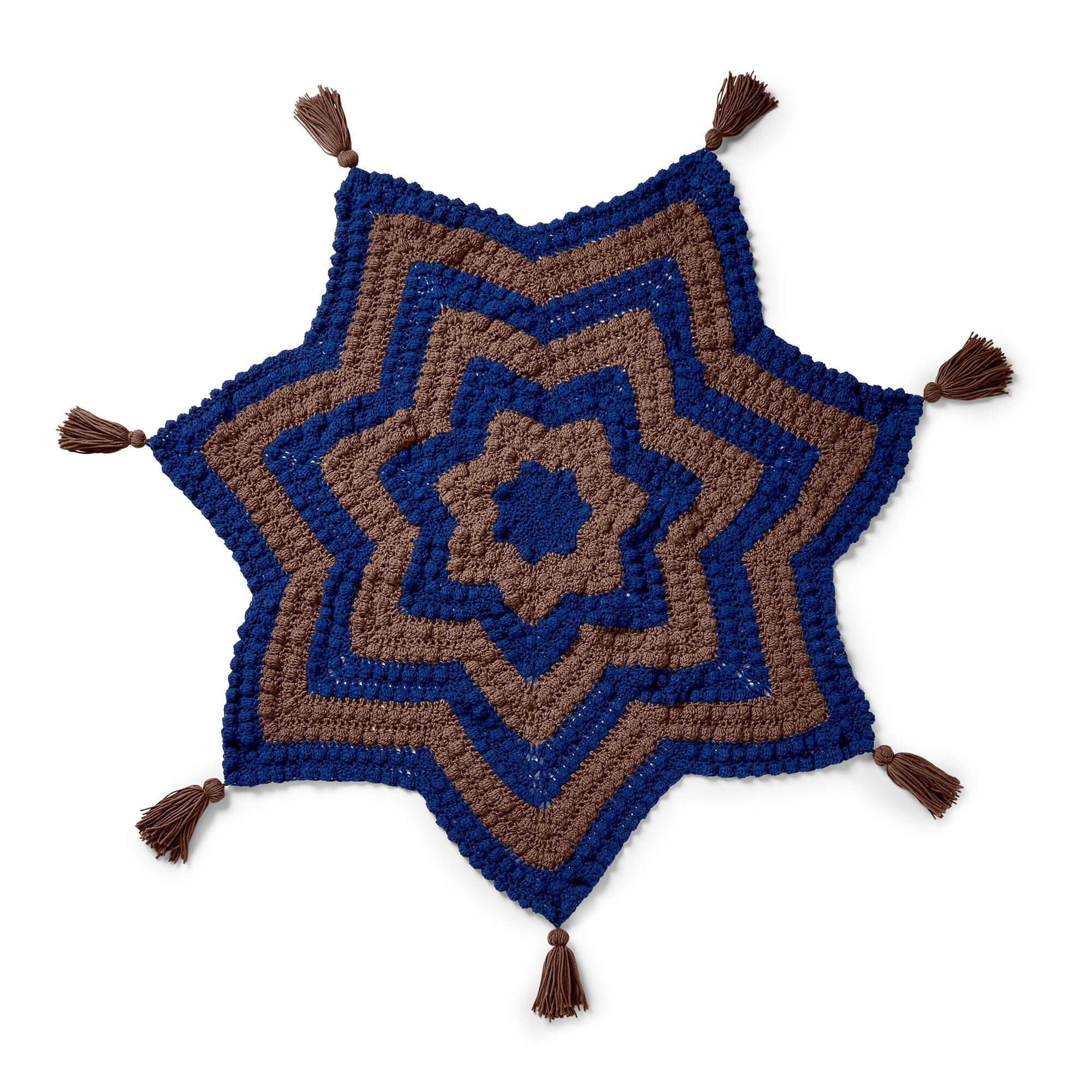}{
   ...\textcolor{blue}{A Midnight Blue} ...
...\textcolor{brown}{B Copper Brown} ...
  \textcolor{cyan}{Ch4. Join with sl st to first ch to
form ring} ... 
    4th rnd: Sl st in next dc. Ch 2. 3 hdc
in same dc as sl st. *\textcolor{magenta}{Bobble} in next
st... \textcolor{purple}{Tassels (make 7)}...
  }
  \caseblock{Gemini}{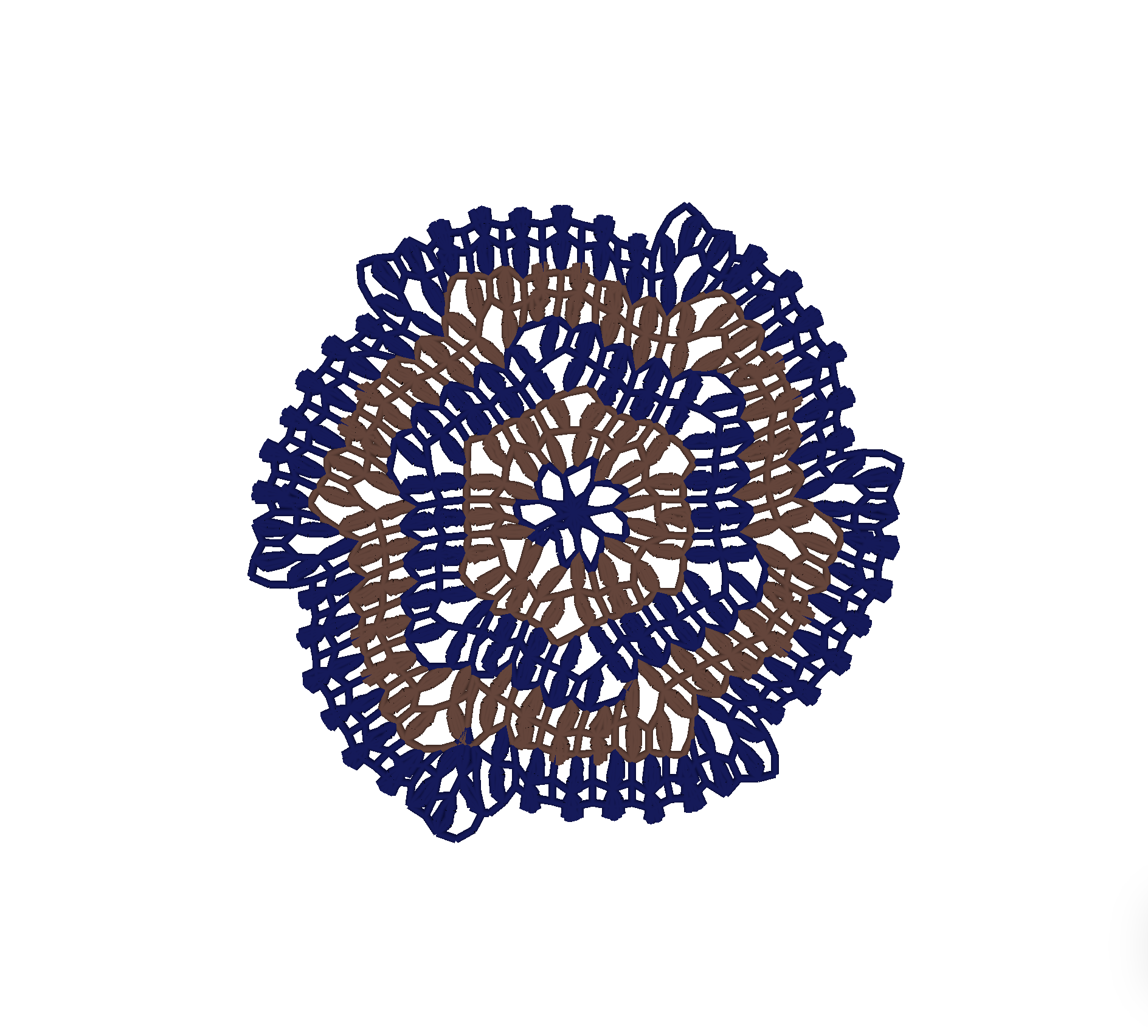}{
  ...
  With \textcolor{blue}{Color A (Blue)}, make a \textcolor{cyan}{magic ring}
  ... Join \textcolor{brown}{Color B (Brown)}
  ...
 Rnd 4:Sl st into ch-2 sp. Ch 1, (\textcolor{magenta}{Bobble}
    ...
    Make \textcolor{purple}{7 tassels} with Color B and attach one to each ch-2 point of the \textcolor{orange}{star}...
  }
  \caseblock{GPT-4o}{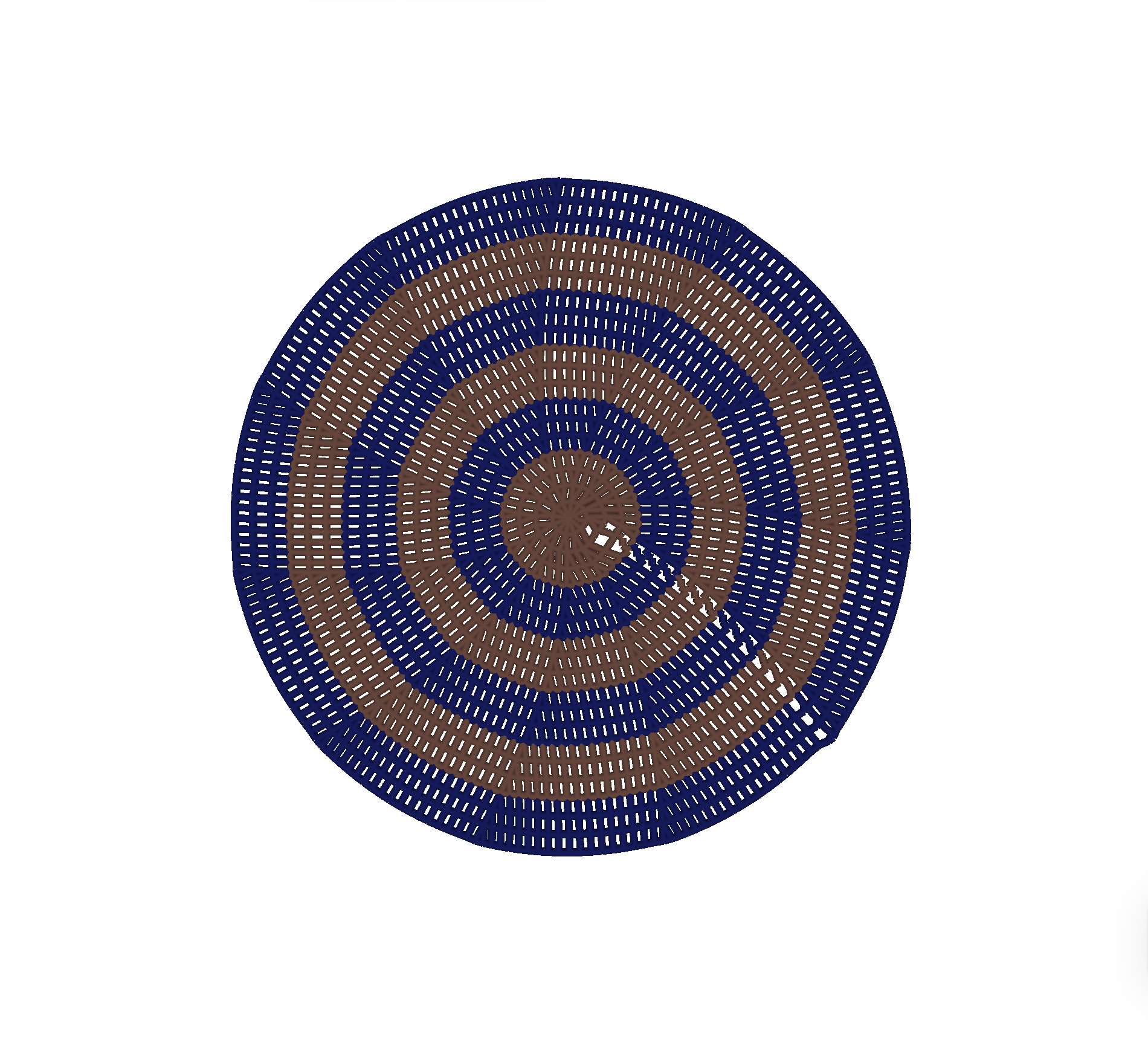}{
   ... \textcolor{orange}{Star-Shaped Crochet Blanket Pattern} ... Yarn: \textcolor{brown}{Brown} and \textcolor{blue}{Blue}...
   \textcolor{red}{With Brown yarn}, make a \textcolor{cyan}{magic ring}.
   ...
   \textcolor{purple}{Tassels}: \textcolor{red}{Cut 4} lengths of Brown yarn ...
}
  \caseblock{Claude}{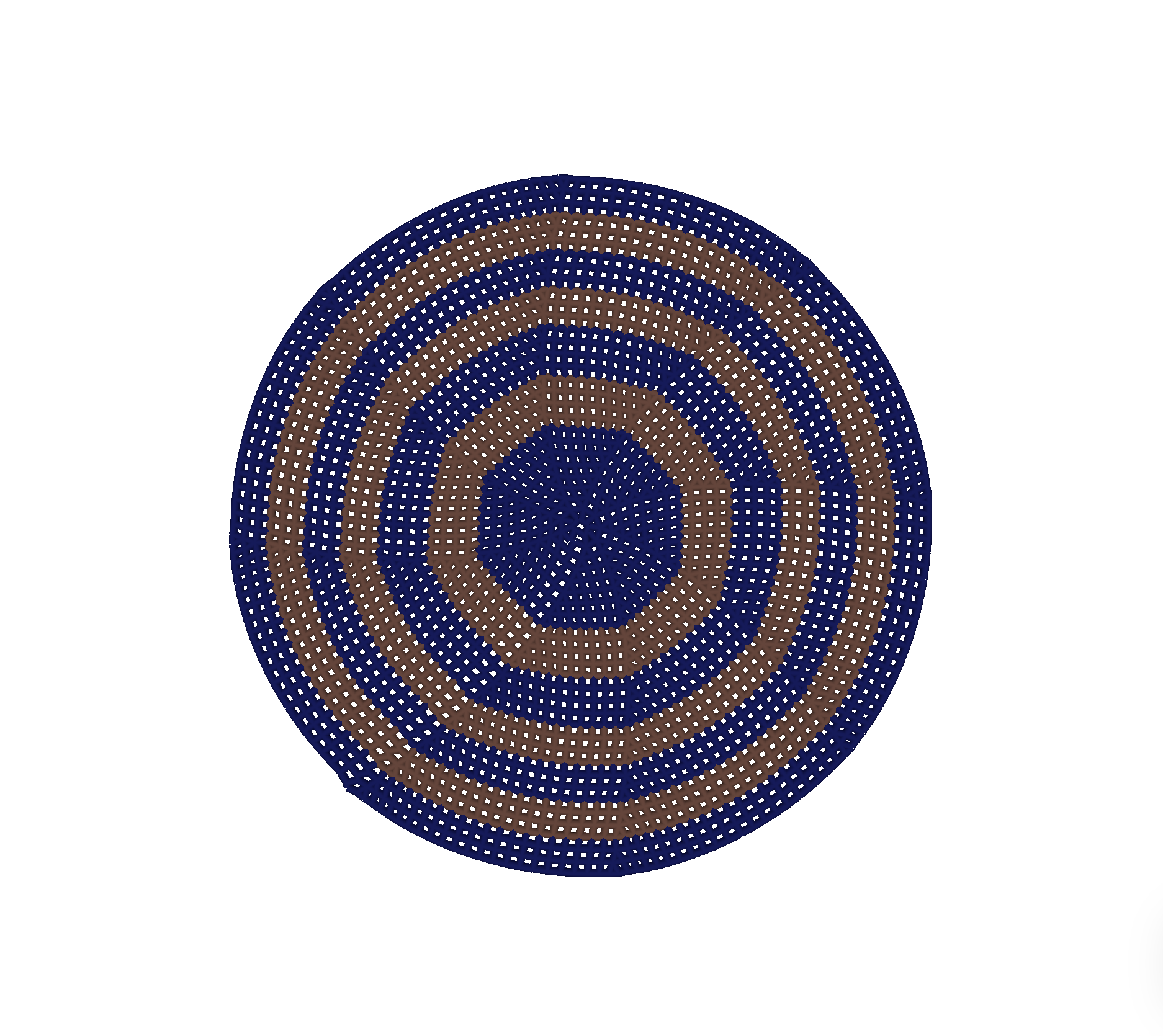}{
  ...
  \textcolor{red}{8-Point} \textcolor{orange}{Star Afghan}
  ...
  Worsted weight yarn in \textcolor{blue}{blue} and \textcolor{brown}{brown}
  ...
With blue, \textcolor{cyan}{ch 4, join with sl st to form ring}.
...
\textcolor{purple}{Tassels}: Cut 
...
  }
  \caseblock{Qwen-VL-7B}{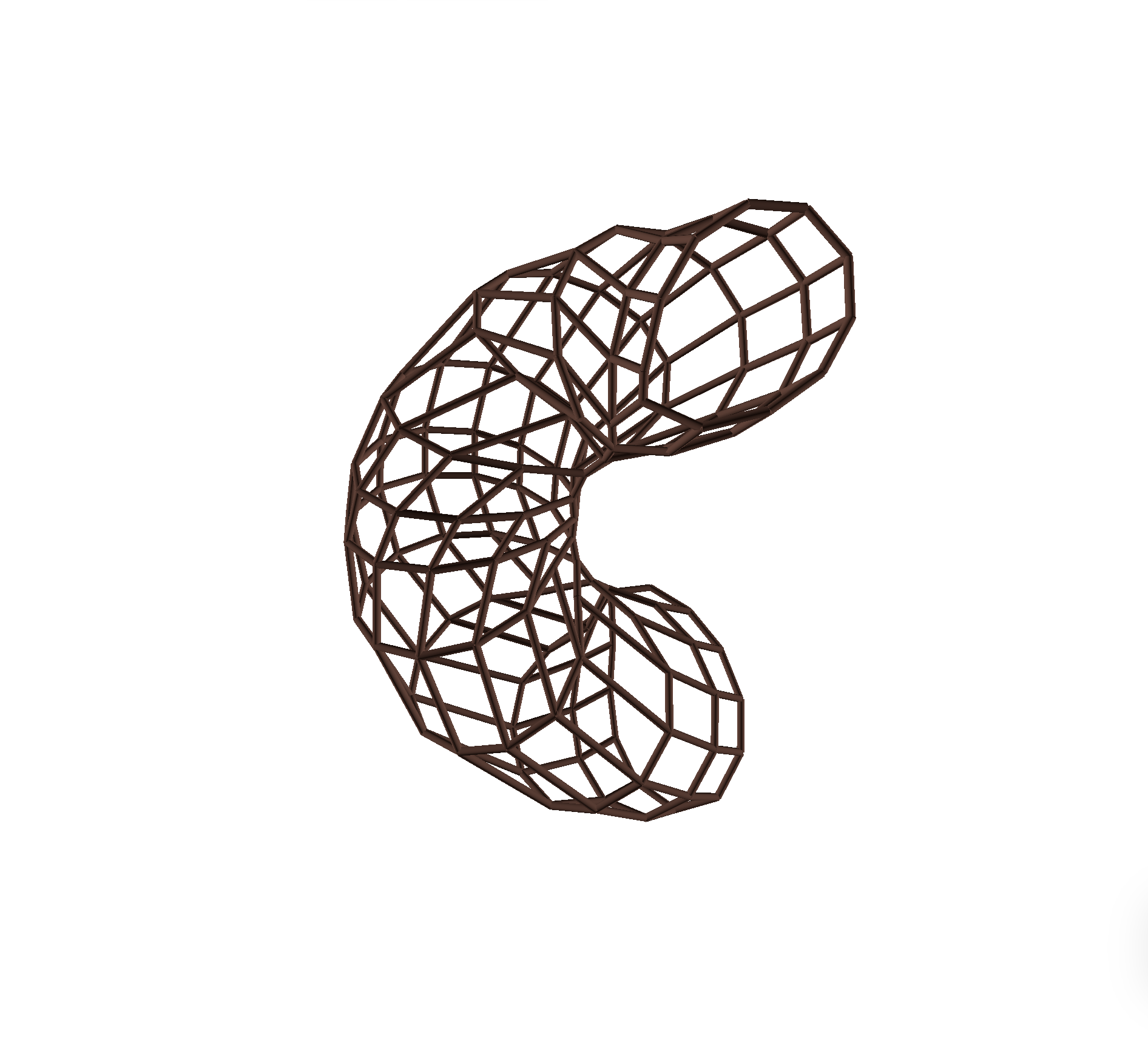}{...
\textcolor{orange}{Star} Motif Crochet Pattern
...
Worsted weight yarn in \textcolor{blue}{navy blue} and \textcolor{brown}{brown}
...
\textcolor{purple}{Tassels}...
Round 1:\textcolor{red}{Ch 10}, join with sl st to form a \textcolor{cyan}{ring}
...
}
  \caseblock{DeepSeek-VL}{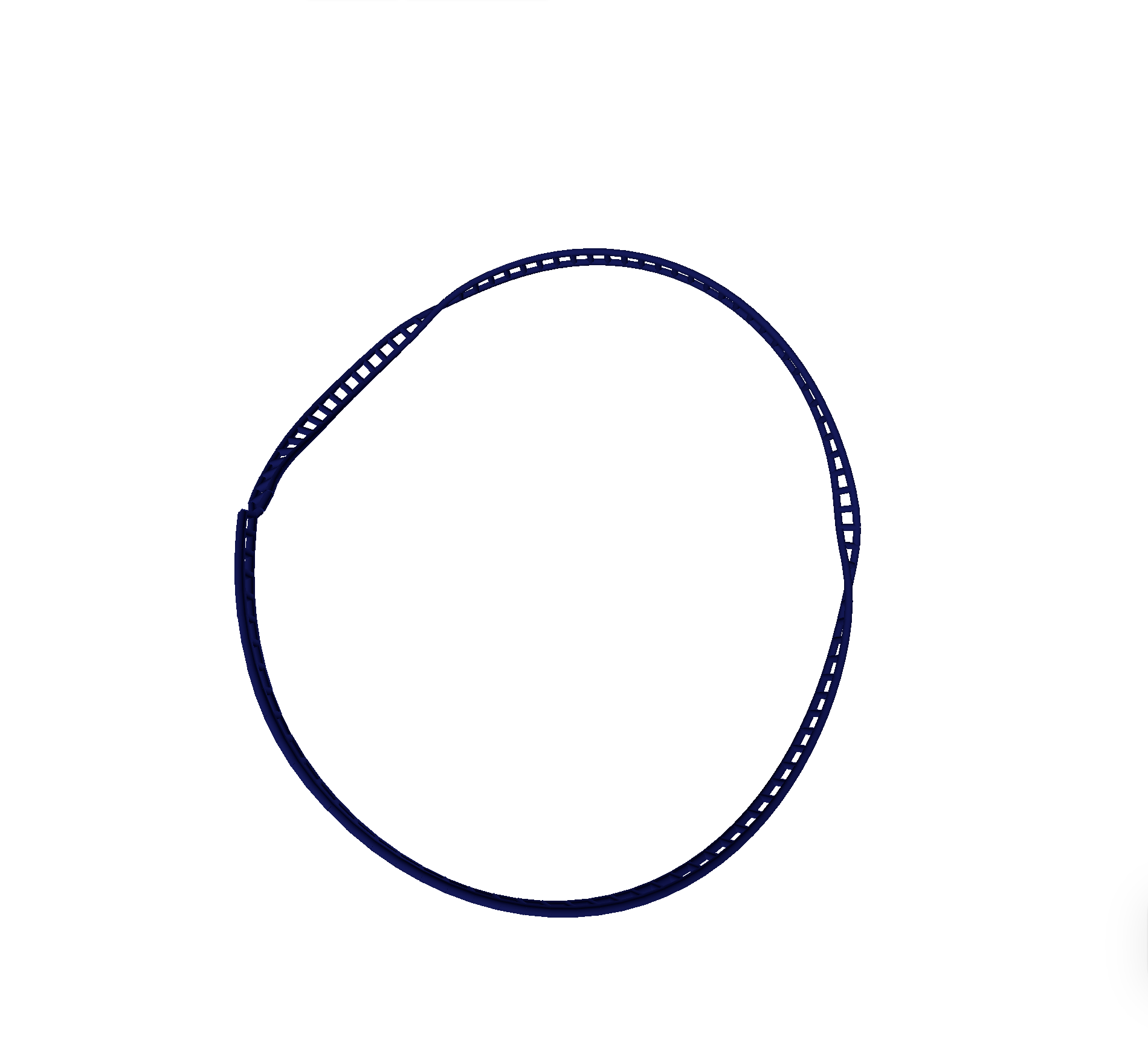}{
    \textcolor{red}{ch 100, sc in 2nd ch from hook, sc in each ch across, join with sl st in 1st sc, 100 loops made.}
  }

  \vspace{2pt}
\caption{
\textbf{Case study for Task~C: Instruction Generation.}
Each row
shows the DSL-rendered output generated from the model’s natural-language
instructions and the color-coded instruction extract below it. Matching colors
denote semantically corresponding elements across the reference and model
outputs, while \textcolor{red}{red} marks incorrect or hallucinated steps.
The ground truth is a seven-point star with alternating blue and brown yarn and
tassels attached at each point. Gemini and GPT-4o generate structured and
mostly coherent instructions but misconstruct the global geometry, producing a
circular motif rather than a star. Claude and Qwen2-VL-7B misinterpret the
shape more severely, producing circular or distorted wireframe-like forms.
DeepSeek-VL collapses entirely into a degenerate single-loop pattern. Gemini is
the only model to explicitly recognize the motif as a seven-point star, but its
instructions still fail to produce the correct star topology. To ensure transparency, we include the fully generated instructions for all models in Appendix~\ref{app:claude_example}.}
  \label{fig:case_study_main}
\end{figure*}

\section{Experiments}

We evaluate a representative set of widely used vision–language models spanning open and closed ecosystems. For open source models, we include BLIP-2 Flan-T5 XL \citep{blip}, Google Gemma~3 (4B and 27B) \citep{team2024gemma}, DeepSeek-VL~7B \citep{lu2024deepseek}, and Qwen2-VL (7B and 72B) \citep{wang2024qwen2}, covering a range of architectures and parameter scales. For closed source models, we evaluate GPT-4o \citep{hurst2024gpt} , Gemini~2.5 Flash-Lite \citep{comanici2025gemini}, and Claude Sonnet~4 \citep{anthropic2025claude_sonnet4}, which represent the strongest publicly accessible multimodal systems. These models span diverse architectures and parameter scales, providing a diverse and meaningful basis for assessing current multimodal capabilities on perception, retrieval, and procedural reasoning tasks.

We evaluate all models on the four CrochetBench tasks described in Section~\ref{sec:tasks}. For Tasks A–C and the project-level setting of Task D, we adopt a stratified 80/5/15 train/validation/test split and report all results on the held-out test set. To avoid cross-task leakage, splits are constructed at the pattern level and shared across tasks. The step-level variant of Task D is evaluated on a curated set of 119 instances, treated as a standalone diagnostic test set. The prompts used for evaluation are provided in Appendix~\ref{appendix:prompts}. Results and analysis are listed below.

\paragraph{Perception and grounding improve with scale, but procedural generation collapses.}
Table~\ref{tab:combined_all} summarizes results across Stitch Recognition (Task~A), Instruction Selection (Task~B), and Instruction Generation (Task~C). Closed-source models achieve the strongest recognition performance, with Claude Sonnet~4 obtaining the highest F1 score (60.94\%), and DeepSeek-VL~7B leading among open models (60.60\%). A clear precision--recall trade-off emerges across model scales: smaller models such as DeepSeek-VL~7B and Qwen2-VL~7B  achieve higher recall but lower precision, whereas larger models such as Qwen2-VL~72B and closed-source models favor precision over recall. This pattern reflects different grounding strategies: smaller models tend to broadly enumerate common stitch types to maximize coverage, while larger models adopt a more selective, conservative approach, yielding higher precision but incomplete visual grounding.

Although larger models capture more fine-grained visual cues, accuracy remains far from saturated, and Instruction Selection shows similarly limited progress, with most large and closed-source models clustering around 55--60\% accuracy,  indicating that visual–textual alignment still depends on shallow correlations rather than robust grounding. This observation is consistent with prior work \citep{zhang2024visually}, which shows that visually grounded language models often underperform specialized vision models on fundamental perception tasks such as image classification, suggesting that their visual representations are not reliably grounded despite large model capacity.

 These limitations become more pronounced in Task~C. Natural-language instruction generation remains extremely challenging for all model, with BLEU, ROUGE-L, and ChrF scores uniformly low; even the strongest system, Gemini 2.5 Flash-Lite, achieves only 4.93\% BLEU and 30.50\% ChrF. The sharp drop from Tasks~A and B to~C shows that models capable of recognizing stitches or retrieving plausible text still fail to synthesize coherent multi-step procedures, reflecting fundamental gaps in procedural reasoning, symbolic consistency, and pattern-structure understanding.

\paragraph{Surface-level fluency does not imply procedural correctness.} 
Figure~\ref{fig:case_study_main} presents a case study comparing model-generated natural-language instructions with their corresponding DSL renderings. Qwen2-VL-7B and DeepSeek-VL collapse into non-star geometries, revealing unstable procedural logic. GPT-4o and Claude produce structurally well-formed, human-like crochet instructions, including explicit materials sections and round-level organization, and correctly capture local yarn colors. However, they fundamentally misinterpret the global motif: GPT-4o reconstructs a four-point star and begins with the brown yarn instead of blue, while Claude generates an eight-point motif rather than the intended seven. Gemini most accurately identifies the seven-point structure and selects plausible constructs such as bobbles for the star tips, but structural inconsistencies remain and yield visibly distorted shapes. These examples demonstrate that models can generate fluent, crochet-like descriptions while failing to preserve the algorithmic structure required for faithful pattern synthesis.

\begin{figure}[H]
    \centering
    \includegraphics[width=\linewidth]{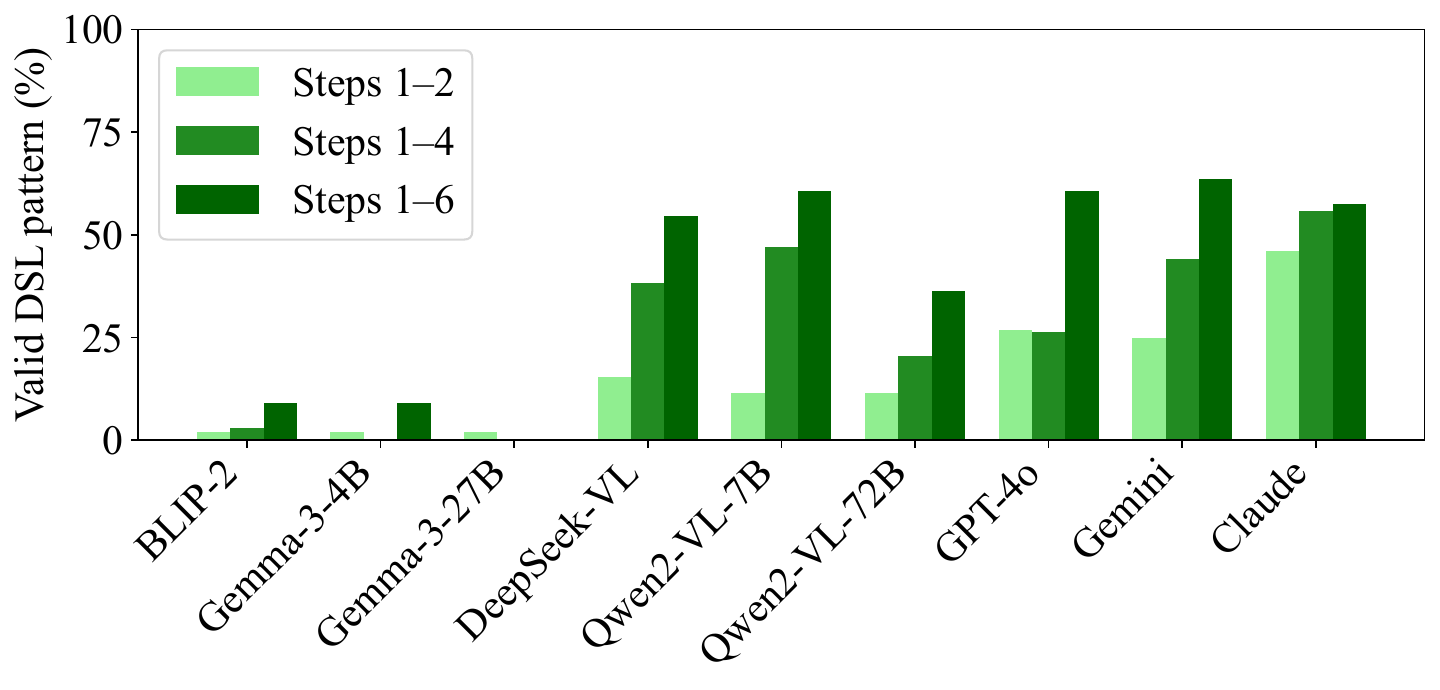}
\caption{
\textbf{Task~D step-level translation}
  results showing the proportion of generated DSL lines that successfully compile for early (Steps~1--2), middle (Steps~1--4), and late (Steps~1--6) stages of crochet patterns. 
 Across all models, valid pattern rates increase as more context is provided, but overall accuracy remains low. Larger models (e.g., Qwen2-VL-72B and Gemma-3-27B) do not consistently outperform their smaller counterparts, highlighting scale alone does not improve program-level structural reasoning.
}
\label{fig:step_level_bar}
\end{figure}
\paragraph{Supervised finetuning improves fluency but not procedural correctness.}
We finetune Qwen2-VL-7B-Instruct using standard training settings (AdamW, 3 epochs, effective batch size 8).
As shown in Table~\ref{tab:combined_all} (bottom line), finetuning substantially improves all generation metrics: BLEU increases from 1.67\% to 5.64\% (+238\%), ROUGE-L from 21.10\% to 25.10\% (+19\%), and ChrF from 15.99\% to 22.39\% (+40\%). The finetuned 7B model outperforms all evaluated closed-source systems, including GPT-4o, Claude Sonnet~4, and Gemini~2.5 Flash-Lite. Despite these gains, absolute performance remains low (5.64\% BLEU), indicating that finetuning does not resolve core procedural failures. Qualitative analysis shows that improvements are largely superficial: outputs exhibit more consistent formatting, improved stitch vocabulary, and better adherence to crochet conventions, but still contain frequent stitch-count errors, malformed repeats, and incorrect global structure (e.g., rounds vs. rows, increases/decreases). These findings suggest that finetuning primarily enhances surface fluency rather than inducing robust algorithmic or structural reasoning.

\paragraph{Early-step instability reveals limits of procedural reasoning.}
Figure~\ref{fig:step_level_bar} shows step-level results on Task~D. Valid Pattern Rate increases with pattern depth but remains low overall: most models achieve under 15\% validity in the first two steps, improve modestly in steps~3–4, and reach only 55–65\% in later steps. This pattern reflects the difficulty of the initial steps, which must correctly initialize the program state such as defining stitch variables and maintaining balanced grouping. Errors made early propagate irreversibly, and later correctness often occurs only when the initial state is accidentally valid, indicating reliance on continuation heuristics rather than genuine procedural understanding. Larger models do not consistently perform better: Qwen2-VL-72B underperforms Qwen2-VL-7B, and Gemma-3-27B underperforms Gemma-3-4B, suggesting that increased capacity improves descriptive fluency more readily than symbolic stability, and that scaling alone is insufficient for grammar-sensitive procedural tasks.

\begin{figure}[H]
    \centering
    \includegraphics[width=\linewidth]{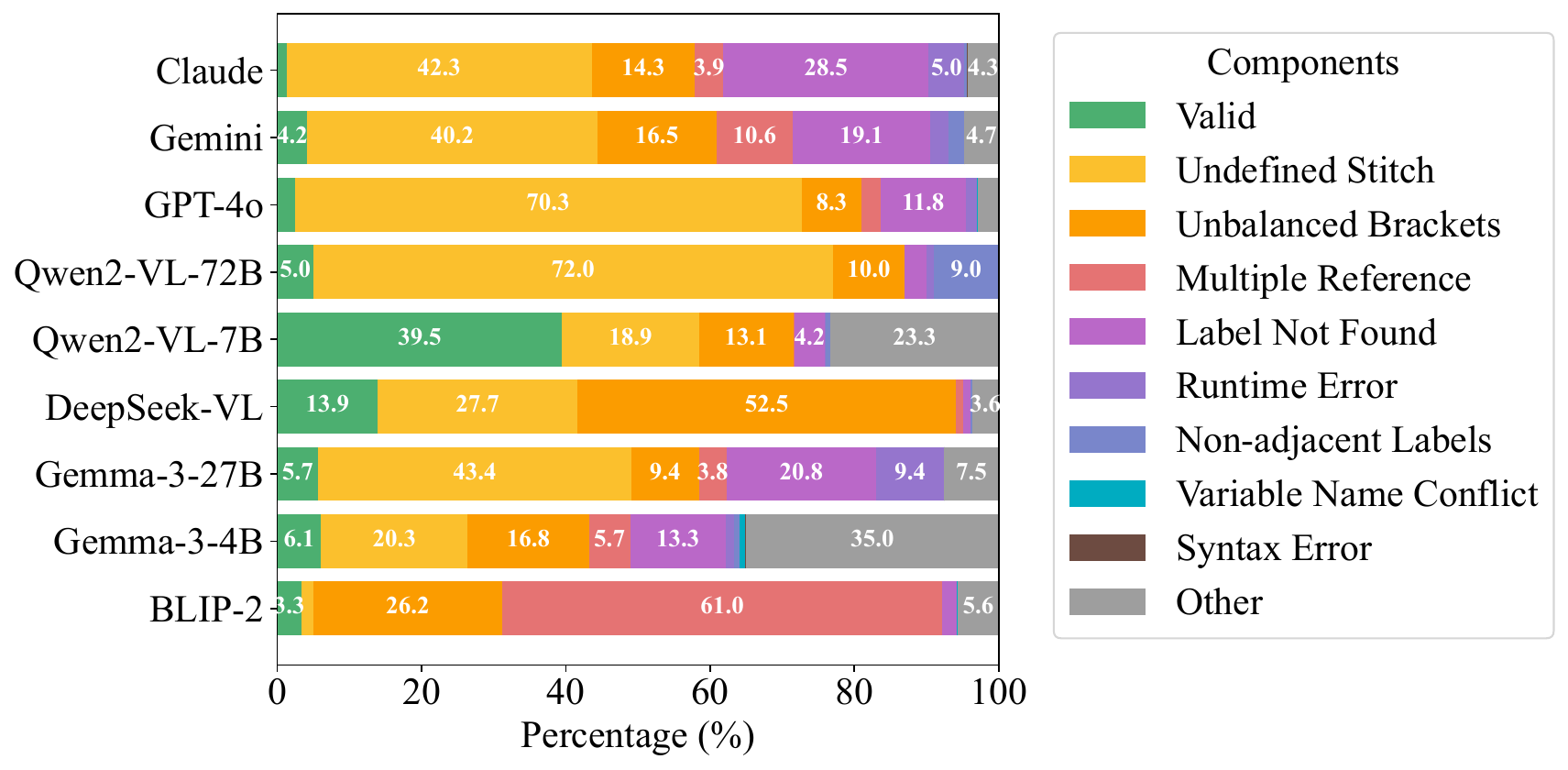}
\caption{
\textbf{Distribution of project-level DSL translation outcomes for each model, broken down into valid outputs and error categories.} Across all models, invalid programs dominate, with most failures arising from undefined stitches, unbalanced brackets, and multiple-reference errors. The spread of error types further illustrates the difficulty of maintaining global consistency and symbolic correctness when generating full crochet programs.}
\label{fig:project}
\end{figure}
\paragraph{Project-level DSL synthesis reveals severe long-range structural failures and does not improve with model scale.}
Figure~\ref{fig:project} highlights the fragility of full-program CrochetPARADE synthesis. Valid outputs are rare: even the strongest models (Claude, Gemini, and GPT-4o) produce only 5--8\% executable programs. Interestingly, smaller open-source models such as Qwen2-VL-7B and DeepSeek-VL achieve higher valid rates. However, qualitative inspection reveals that these gains are largely driven by copying or closely following the few-shot exemplars, rather than genuine procedural reasoning.

Across models, the most frequent failure modes (undefined stitches and unbalanced brackets) indicate poor control over the DSL vocabulary and grouping structure. Additional errors, including multiple references, non-adjacent labels, and runtime failures, further reveal difficulties in maintaining consistent state and long-range structural dependencies across an entire pattern. Notably, scaling does not alleviate these issues and can even exacerbate them. Larger models exhibit higher rates of undefined-stitch errors, suggesting a tendency toward uncontrolled symbolic invention (e.g., Qwen2-VL-72B: 72.0\% vs.\ Qwen2-VL-7B: 18.9\%; Gemma-3-27B: 42.4\% vs.\ Gemma-3-4B: 20.3\%).

\begin{figure}[H]
    \centering
    \includegraphics[width=\linewidth]{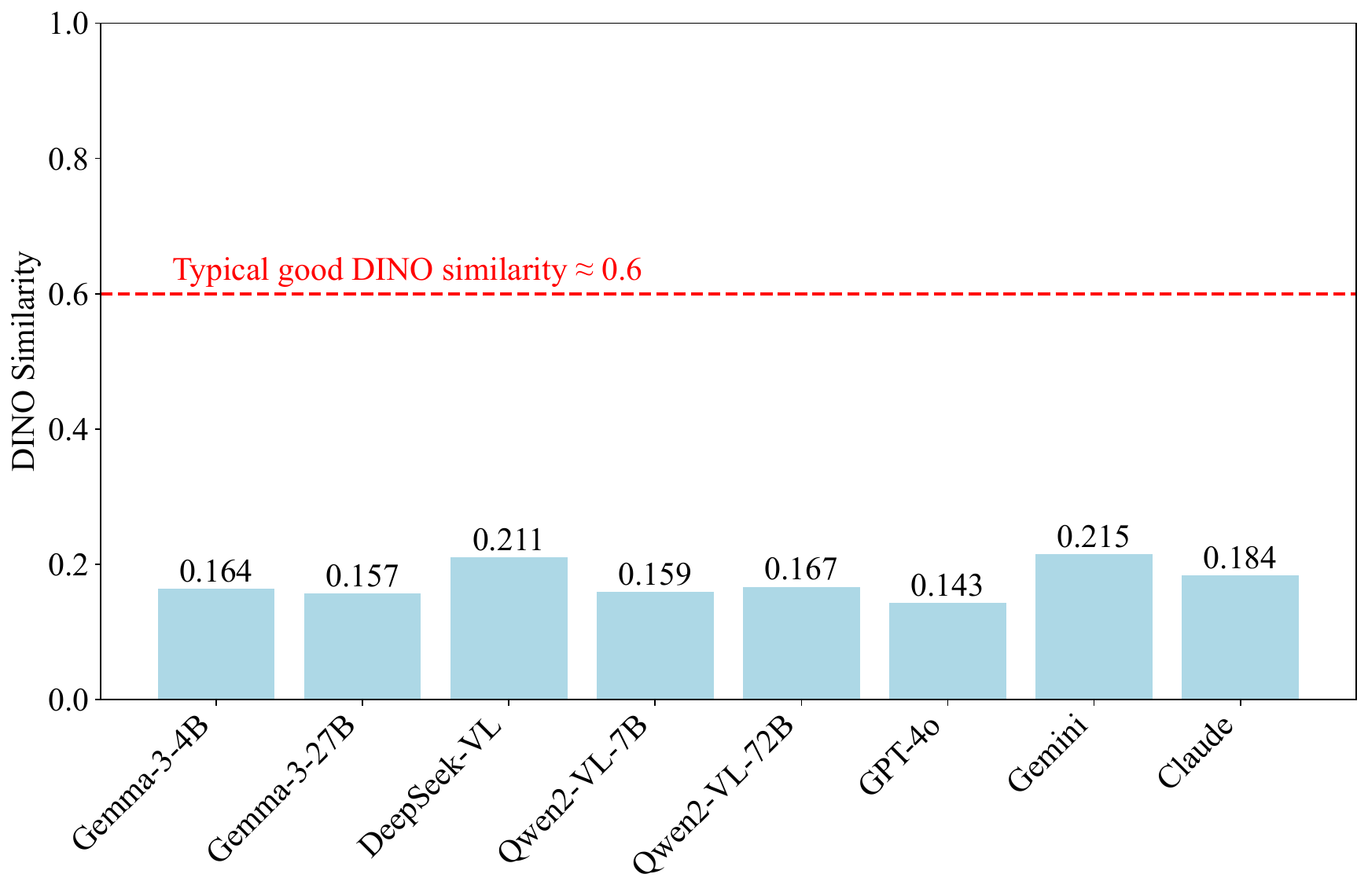}
\caption{
\textbf{Task~D Project-level translation}
results evaluating with \textbf{DINO similarity} between ground-truth images and DSL-rendered outputs generated from each model’s DSL program (valid executable portion only). 
The red line marks an approximate ``good'' similarity threshold. 
All models fall well below this level, indicating that even executable DSL programs rarely reproduce the correct crochet pattern.}
\label{fig:dino}
\end{figure}

\paragraph{Image-based similarity confirms lack of global structural fidelity.}
Compilation verifies syntactic and structural correctness but cannot determine whether two DSL programs are semantically equivalent. To address this gap, we compute DINO similarity between the target crochet product image and the rendering produced from each model’s executable program (valid outputs only). Figure~\ref{fig:dino} shows that similarity scores remain uniformly low across all models (0.14--0.22), far below the typical threshold (0.6) for visually matched crochet images. Even when a model produces a compilable DSL program, the resulting rendering generally bears little resemblance to the intended pattern, indicating that syntactic validity does not imply correct procedural structure. The consistently low similarities reinforce that current multimodal LLMs fail to capture the global geometry and layout required for visually faithful crochet synthesis.

%% file: section/related_work.tex
\section{Related Work}

Multimodal benchmarks have traditionally emphasized descriptive image--text alignment, as in COCO \citep{lin2015microsoftcococommonobjects} and Flickr30k \citep{plummer2016flickr30kentitiescollectingregiontophrase}. More recent datasets extend to instructional or procedural understanding, including Recipe1M+ \citep{marin2019recipe1mdatasetlearningcrossmodal} and large-scale instructional video corpora such as YouCook2 and HowTo100M \citep{zhou2017automaticlearningproceduresweb, miech2019howto100mlearningtextvideoembedding}. However, these benchmarks primarily evaluate semantic alignment, retrieval, or temporal recognition, and do not assess whether a model can generate or execute a \emph{correct} procedure.


Texile crafts provide precisely such a domain. Crochet patterns specify symbolic, stepwise 
procedures that determine the topology and geometry of a final physical artifact. Prior work in 
this area including Digital Crochet \citep{digital_crochet} and Neural Inverse Knitting 
\citep{kaspar2019neuralinverseknittingimages} demonstrates the feasibility of representing 
textile structures in machine-readable form but remains limited in scale and modality. 
CrochetBench builds on this emerging direction by providing thousands of real crochet patterns 
with paired images and natural-language instructions.

To evaluate procedural correctness, CrochetBench adopts an executable domain-specific language 
(CrochetPARADE), linking our tasks to program synthesis benchmarks such as HumanEval 
\citep{chen2021evaluatinglargelanguagemodels}, MBPP \citep{austin2021programsynthesislargelanguage}, 
and Spider \citep{yu2019spiderlargescalehumanlabeleddataset}. In multimodal settings, 
image-to-program benchmarks such as Im2LaTeX-100K \citep{deng2017imagetomarkupgenerationcoarsetofineattention} 
and pix2code \citep{beltramelli2017pix2codegeneratingcodegraphical} similarly leverage executable 
formalisms for rendering-based evaluation. CrochetBench extends this executable perspective to 
textile crafts: patterns compile to structured instructions that can be rendered and validated, 
providing functional evaluation that tests whether a model’s output \emph{actually works}. This offers a lightweight alternative to domains such as chemistry or cooking , 
where validating a procedure requires physical or chemical experiments \citep{huang2025chemhgnn} that are slow, 
costly, or impractical to scale.

%% file: appendix/crochet_primer.tex
\appendix
\section{Crochet Primer}
\label{app:crochet-primer}

Crochet patterns describe how to construct a textile artifact through a sequence of 
symbolic stitch instructions. Each instruction specifies an operation performed with a 
hook and yarn, and the resulting pattern is defined by the order, repetition, and 
spatial arrangement of these stitches. This appendix summarizes only the conventions 
needed to interpret the examples in our benchmark.

\paragraph{Basic stitch types.}
Crochet relies on a small vocabulary of atomic stitches, each producing a loop with a 
characteristic height and structure. The most common stitches in U.S.\ notation are:
\begin{itemize}
    \item \textbf{ch} (chain): foundational stitch used to begin rows or rounds.
    \item \textbf{sc} (single crochet): a short, dense stitch.
    \item \textbf{hdc} (half double crochet) and \textbf{dc} (double crochet): taller stitches 
    that build height more quickly.
    \item \textbf{sl st} (slip stitch): a joining stitch used for connecting motifs or closing rounds.
\item \textbf{bobble} (bobble stitch): a cluster of 5 partially completed double crochet stitches closed together into a single stitch.
\end{itemize}
These stitches can be combined in rows (worked back and forth) or rounds (worked in a circle).

\paragraph{Pattern syntax and structure.}
Crochet instructions follow a compact symbolic notation. A pattern is organized into 
\emph{rows} or \emph{rounds}, each specifying a sequence of stitches. For example:
\[
\texttt{Row 3: Ch 1, sc in each st across, turn.}
\]
Instructions may include:
\begin{itemize}
    \item \textbf{Repetition}: indicated by parentheses and a multiplier, e.g.,
    \texttt{(sc, ch 1) 3 times}.
    \item \textbf{Increases/decreases}: e.g., \texttt{2 sc in next st} (increase) or 
    \texttt{sc2tog} (single-crochet two stitches together; decrease).
    \item \textbf{Stitch counts}: patterns often end rows or rounds with “—\textit{N} sc,” 
    indicating the number of stitches that should remain.
\end{itemize}

\paragraph{Relationship to symbolic representations.}
Each crochet instruction corresponds to a local modification of the fabric’s topology. 
This makes crochet patterns naturally suited to symbolic or program-like representations 
such as CrochetPARADE, which encode stitches as structured primitives with explicit 
control flow (loops, groups, labels). Because stitch sequences fully determine the 
geometry of the final artifact, correctness can be assessed by verifying the structure of 
the generated program or by rendering the corresponding stitch graph.

This primer covers the minimal terminology required to interpret our dataset and 
evaluation tasks. For readers interested in additional background, standard crochet 
references provide extended stitch catalogs and diagram conventions.

%% file: appendix/crochetBench_analysis.tex
\section{Additional Dataset Statistics}
\label{app:data-statistics}
\begin{figure}[t]
  \centering
  \vspace{-1em} 
  \includegraphics[width=0.4\textwidth]{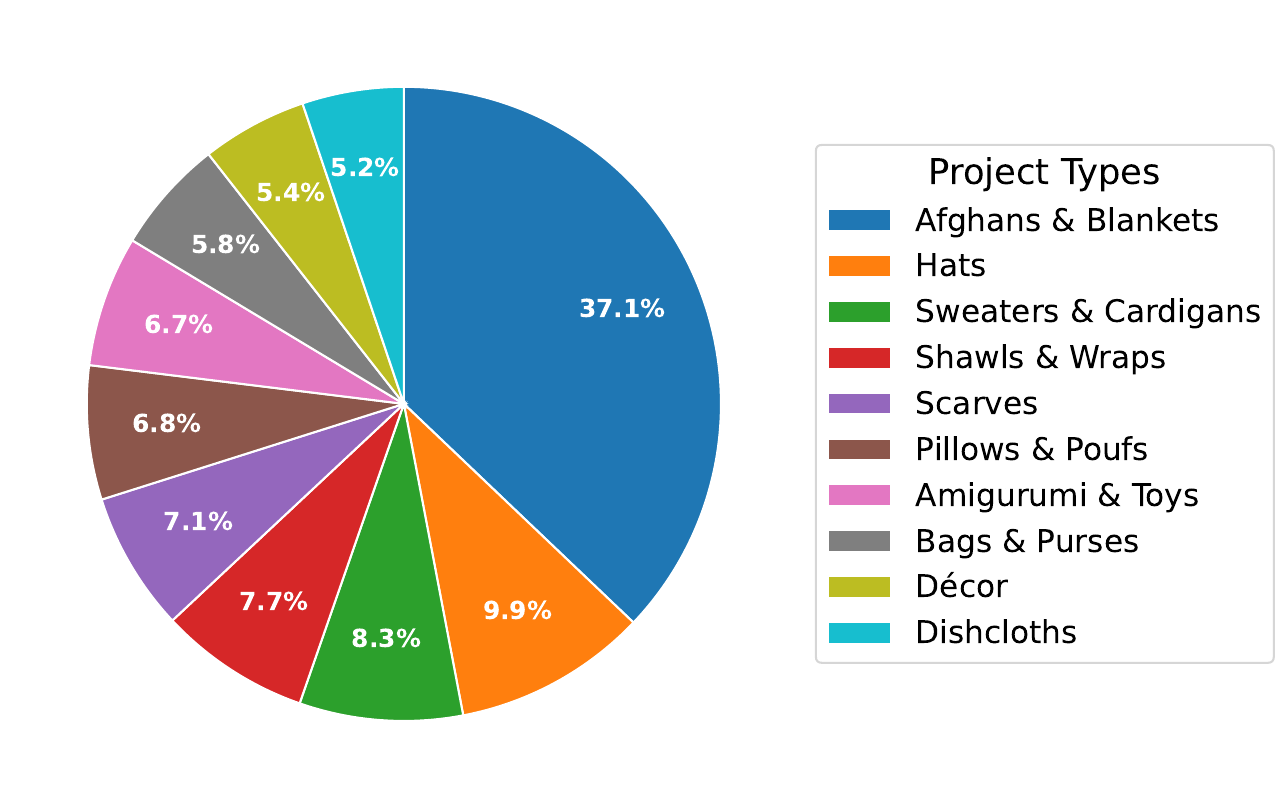}
  \vspace{-1em} 
  \caption{Distribution of the top-10 most common project types in CrochetBench.}
  \label{fig:project-types}
\end{figure}

 \textbf{CrochetBench} supports diverse real-world crochet practices, with project types ranging from simple accessories to complex garments. Figure \ref{fig:project-types} lists the ten most common categories by frequency. The majority of patterns belong to a small number of dominant types—Afghans and Blankets alone account for over one-quarter of the dataset.

\begin{figure}[t]
  \centering
  \vspace{-1em} 
  \includegraphics[width=0.4\textwidth]{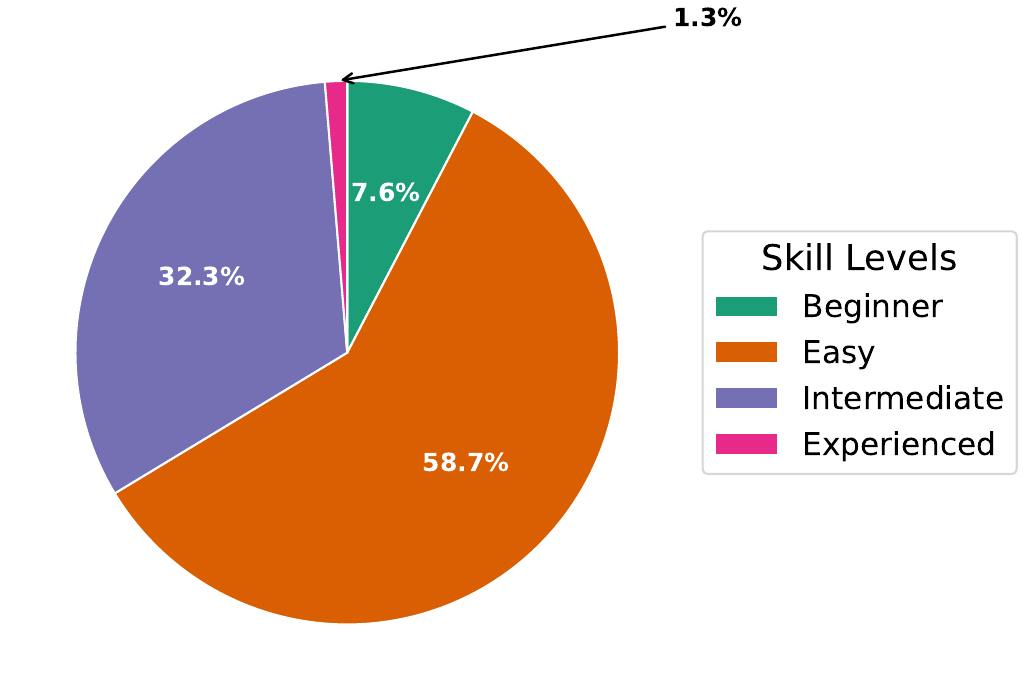}
  \vspace{-1em} 
  \caption{Skill level distribution across the CrochetBench dataset. 
 }
  \label{fig:skill-distribution}
\end{figure}

 Each pattern is labeled with one of four primary skill levels, including \textit{beginner}, \textit{easy}, \textit{intermediate}, or \textit{experienced}. This allows for stratified evaluation across complexity tiers. Figure \ref{fig:skill-distribution} shows the skill level distribution, which is strongly skewed toward beginner-friendly content. Instructional complexity varies substantially across patterns. The number of characters in each instruction ranges from 20 to over 30,000, with a mean of 3,216 and a median of 2,453. Abbreviation counts (i.e., unique stitch tokens per pattern) range from 1 to 31, with an average of 10.6. 

 We observe a clear correlation between skill level and instruction length: beginner patterns tend to be short and use fewer abbreviations, while experienced patterns are significantly longer and more symbolically dense.

 In addition to symbolic complexity, the dataset contains 3,143 abbreviation instances
 mapped to 789 unique standardized stitch tokens. This lexical mapping enables tasks such as vocabulary translation, sequence generation, and instruction validation. Beyond raw instructions,
the structured schema also records rich metadata. A representative dataset entry is provided in Table~\ref{app:dataset-example}.


\begin{table*}[htbp]
\centering
\begin{tabular}{p{2.5cm} p{12.5cm}}
\toprule
\textbf{Field} & \textbf{Value} \\
\midrule
Pattern Name & SKULL TRICK OR TREAT BAG (TO CROCHET) \\
Skill Level & Intermediate \\
Project Type & Bags or Purses \\
Measurements & 15 cm diameter $\times$ 15 cm high (excluding handle) \\
Gauge & 13 sc and 14 rows = 10 cm \\
Materials & Lily\textsuperscript{\textregistered} Sugar’n Cream (White, Black), 5 mm hook, cardboard \\
Image & \url{https://www.yarnspirations.com/cdn/shop/products/SCC0303-005314M.jpg} \\
Source & \texttt{input\_file/Bags+Purses/SCC0303-005314M.pdf} \\
Instructions & \textbf{Instructions:}\par
Note: Ch 2 at beg of each rnd counts as hdc.\par
\par
\textbf{BAG}\par
With MC, ch 4. Join with sl st to form ring.\par
1st rnd: Ch 2. 11 hdc in ring. Join with sl st to top of ch 2. 12 hdc.\par
2nd rnd: Ch 2. 1 hdc in same sp as sl st. 2 hdc in each hdc around. Join. 24 hdc.\par
3rd rnd: Ch 2. 1 hdc in same sp. 1 hdc in next hdc. *2 hdc in next hdc, 1 hdc in next.* Rep around. Join. 36 hdc.\par
4th rnd: Ch 2. 1 hdc in each hdc around. Join.\par
5th rnd: Ch 2. 1 hdc in same sp. 1 hdc in next 2 hdc. *2 hdc, 1 hdc in next 2.* Join. 48 hdc.\par
6th rnd: As 4th rnd.\par
7th rnd: Ch 2. 1 hdc in next 2 hdc. *2 hdc, 1 hdc in next 3.* Rep. 60 hdc.\par
8th rnd: Ch 2. Back loops only, 1 hdc around. Join.\par
9th--13th rnds: Ch 2. 1 hdc in each hdc around. Join.\par
14th rnd: Ch 2. 1 hdc in same sp. 1 hdc in next 4 hdc. *2 hdc, 1 hdc in next 4.* Join. 72 hdc.\par
15th rnd: Ch 2. 1 hdc in same sp. 1 hdc in next 5 hdc. *2 hdc, 1 hdc in next 5.* Join. 84 hdc.\par
16th--22nd rnds: Ch 2. 1 hdc in each hdc around. Join.\par
23rd rnd: Ch 2. 1 hdc in next 4 hdc. *Hdc2tog, 1 hdc in next 5.* Rep. Hdc2tog. Join. 72 sts.\par
24th rnd: Ch 2. 1 hdc in next 3 hdc. *Hdc2tog, 1 hdc in next 4.* Rep. Join. 60 sts.\par
25th rnd: Ch 2. 1 hdc in next 2 hdc. *Hdc2tog, 1 hdc in next 3.* Rep. Join. 48 sts. Fasten off.\par
\par
\textbf{Eyes (Make 2)}\par
With A, ch 8.\par
1st rnd: 2 sc in 2nd ch from hook. 1 sc in next 5 ch. 3 sc in last ch. Continue on rem loops, 1 sc in each ch. Join. 17 sc.\par
2nd rnd: Ch 1. 3 sc in first sc. 1 sc in next 7 sc. 3 sc in next sc. 1 sc in next 8 sc. Join. Fasten off.\par
\par
\textbf{Handle}\par
With MC, ch 45.\par
1st row: 1 sc in 2nd ch from hook. 1 sc across. 44 sc. Turn.\par
2nd row: Ch 1. 1 sc across. Turn.\par
Rep last row 4 more times. Fasten off.\par
\par
\textbf{Finishing}\par
Sew Eyes to Bag. Embroider mouth and teeth with A. Attach Handle. Cut cardboard circle to fit bottom.\par
\\
\bottomrule
\end{tabular}
\caption{Representative pattern entry from CrochetBench.}
\label{app:dataset-example}
\end{table*}

\subsection{Skill Level Distribution}

Table~\ref{tab:overall-skill} and Figure~\ref{fig:skill-distribution} summarizes the overall distribution of skill levels across the CrochetBench dataset. The majority of patterns are labeled as \emph{easy} (58.7\%), followed by \emph{intermediate} (32.3\%). Only a small fraction are classified as \emph{beginner} (7.6\%) or \emph{experienced} (1.3\%).\footnote{Three additional rare labels were observed: \texttt{easy to intermediate} (1 pattern), \texttt{beginners} (1 pattern), and \texttt{beginner/easy} (1 pattern). Together they account for $<0.1\%$ of the dataset.} One pattern (0.02\%) is missing an annotated skill level.

\begin{table}[H]
\centering

\begin{tabular}{lrr}
\toprule
\textbf{Skill Level} & \textbf{Count} & \textbf{Percentage} \\
\midrule
Easy          & 3569 & 58.66\% \\
Intermediate  & 1967 & 32.33\% \\
Beginner      & 465  & 7.64\%  \\
Experienced   & 80   & 1.31\%  \\
\midrule
\textbf{Total} & 6084 & 100\% \\
\bottomrule
\end{tabular}
\caption{Overall skill level distribution. Percentages are relative to all patterns with annotated skill levels.}
\label{tab:overall-skill}
\end{table}

\subsection{Abbreviation Statistics}
Abbreviations, such as \texttt{sc}, \texttt{dc}, and \texttt{hdc}, are a distinctive element of crochet instructions. Table~\ref{tab:abbr-stats} reports abbreviation counts across all patterns. Most patterns contain about 10 abbreviations, with values ranging from 1 to 31.

\begin{table}[H]
\centering

\begin{tabular}{l r}
\toprule
\textbf{Statistic} & \textbf{Value} \\
\midrule
Average            & 10.6 \\
Median             & 10.0 \\
Min                & 1 \\
Max                & 31 \\
\bottomrule
\end{tabular}
\caption{Abbreviation count statistics.}
\label{tab:abbr-stats}
\end{table}

\subsection{Overall Instruction Length Statistics}
We analyze the distribution of instruction lengths, measured in raw character counts. As shown in Table~\ref{tab:instr-length-stats}, the average instruction length is over 3,200 characters, while the median is substantially lower at 2,453 characters, reflecting a long-tailed distribution. The most complex patterns extend beyond 30,000 characters, while some very short patterns are as small as 20 characters. 
\begin{table}[H]
\centering

\begin{tabular}{l r}
\toprule
\textbf{Statistic} & \textbf{Value} \\
\midrule
Average            & 3216.0 \\
Median             & 2453.0 \\
Min                & 20 \\
Max                & 30634 \\
25th percentile    & 1511.8 \\
75th percentile    & 4136.2 \\
90th percentile    & 6403.9 \\
\bottomrule
\end{tabular}
\caption{Instruction length statistics (in characters).}
\label{tab:instr-length-stats}
\end{table}

\subsubsection{Instruction Complexity by Skill Level}
\label{app:instruction-complexity}
Instruction length correlates with the designated skill level. As shown in Table~\ref{tab:instruction-complexity}, beginner-level patterns average under 2,000 characters, while intermediate patterns extend to over 4,200. Experienced patterns are the longest, averaging 7,689 characters.
\begin{table*}[ht]
\centering
\begin{tabular}{lcccc}
\toprule
\textbf{Skill Level} & \textbf{Avg. Length} & \textbf{Median Length} & \textbf{Avg. Abbr.} & \textbf{Count} \\
\midrule
Beginner             & 1,674 & 1,365 & 9.2  & 465  \\
Easy                 & 2,761 & 2,182 & 10.8 & 3,569 \\
Intermediate         & 4,221 & 3,387 & 10.7 & 1,967 \\
Experienced          & 7,689 & 6,729 & 9.8  & 80   \\
\bottomrule
\end{tabular}
\caption{Instruction complexity by skill level. Length is measured in characters.}
\label{tab:instruction-complexity}
\end{table*}




\subsubsection{Most and Least Complex Project Types}
We identify the most complex and simplest project types by average instruction length. Tables~\ref{tab:complex-types} and \ref{tab:simple-types} list the top 10 categories. Garments such as dresses, vests, pants, and tunics are the most demanding, with average instructions exceeding 5,800 characters. By contrast, smaller accessories such as cowls, washcloths, scarves, and headbands are substantially shorter, typically under 2,000 characters.

\begin{table*}[htbp]
\centering

\begin{tabular}{lrrr}
\toprule
\textbf{Project Type} & \textbf{Avg. Length} & \textbf{Median} & \textbf{Count} \\
\midrule
Dresses                & 6484.9 & 5799.0 & 34 \\
Vests                  & 6032.0 & 5193.5 & 64 \\
Pants                  & 5866.7 & 5409.0 & 11 \\
Tunics                 & 5850.4 & 5832.0 & 29 \\
Sets                   & 5625.5 & 4847.0 & 111 \\
Sweaters \& Cardigans  & 5429.2 & 5113.0 & 357 \\
Amigurumi \& Toys      & 5322.4 & 4505.0 & 286 \\
Jackets                & 5311.9 & 4831.0 & 31 \\
Onesies \& Rompers     & 5263.4 & 5181.0 & 5 \\
Aprons                 & 4467.8 & 4494.0 & 11 \\
\bottomrule
\end{tabular}
\caption{Top 10 most complex project types (by average instruction length).}
\label{tab:complex-types}
\end{table*}

\begin{table*}[htbp]
\centering

\begin{tabular}{lrrr}
\toprule
\textbf{Project Type} & \textbf{Avg. Length} & \textbf{Median} & \textbf{Count} \\
\midrule
Cowls                 & 1288.3 & 956.5  & 154 \\
Washcloths \& Mitts   & 1502.5 & 1420.0 & 28 \\
Scarves               & 1567.3 & 1221.0 & 304 \\
Headbands             & 1617.5 & 1475.5 & 38 \\
Dishcloths            & 1688.4 & 1571.0 & 222 \\
Coasters              & 1750.3 & 1625.0 & 26 \\
Booties               & 1921.9 & 1938.5 & 24 \\
Jewelry               & 1960.3 & 1549.0 & 55 \\
Super Scarves         & 2007.6 & 1213.0 & 13 \\
Tech Accessories      & 2011.1 & 2099.0 & 13 \\
\bottomrule
\end{tabular}
\caption{Top 10 simplest project types (by average instruction length).}
\label{tab:simple-types}
\end{table*}

%% file: appendix/prompt.tex
\section{Prompts}
\label{appendix:prompts}

\subsection{Task A: Stitch Recognition Prompt}
This task evaluates a model’s ability to identify stitches present in a crochet product image.

\begin{tcolorbox}[title=Stitch Recognition Prompt (Rendered Example), breakable,  colback=white, colframe=black, sharp corners=south]
\textbf{SYSTEM PROMPT}
You are a crochet stitch expert. 

Given an image of a crochet product, identify all stitches that appear.

Requirements:

- Use only standard U.S. crochet abbreviations 

  (e.g., sc, hdc, dc, tr, ch, sl st, pop, etc.).
  
- Output must be a comma-separated list of abbreviations.

- Do not include explanations, extra text, or formatting beyond the list.

\vspace{1em} 
\textbf{USER PROMPT}
Look at this crochet product image and list the stitches used.

[Image]
\end{tcolorbox}

\subsection{Task B: Instruction Selection Prompt}
This task evaluates a model’s ability to choose the correct instructions from multiple-choice options.

\begin{tcolorbox}[title=Instruction Selection Prompt (Rendered Example), breakable, colback=white, colframe=black, sharp corners=south]
\textbf{SYSTEM PROMPT}

You are a crochet expert. 
Your task is to determine which of the given options 
(A, B, C, or D) contains the correct crochet instructions 
for the image shown.

\vspace{1em} 
\textbf{USER PROMPT}

Look at this crochet image and choose which option 
best matches the instructions for making it.

[Image]

Options:
$\{$options text$\}$

Choose exactly ONE option. Your answer should be only 
one letter: A, B, C, or D.
\end{tcolorbox}

\subsection{Task C: Instruction Generation Prompt}
This task evaluates a model’s ability to generate complete crochet instructions from an image.

\begin{tcolorbox}[title=Instruction Generation Prompt (Rendered Example), breakable, colback=white, colframe=black, sharp corners=south]
\textbf{SYSTEM PROMPT}

You are a professional crochet pattern writer. 
Examine the image of the finished crochet product carefully. 
Write a complete set of crochet instructions in the standard style used in published patterns.

Requirements:

- Use standard abbreviations: sc (single crochet), hdc (half double crochet), 

  dc (double crochet), tr (treble), ch (chain), sl st (slip stitch), rep (repeat).
  
- Organize the instructions row by row or round by round (e.g., "Rnd 1: ...", "Row 2: ...").

- If color changes are visible in the image, include them in the pattern.

- Keep the instructions concise and precise, as if for experienced crocheters.

- Output only the crochet pattern. Do not add any explanations, commentary, or extra text.

\vspace{1em} 
\textbf{USER PROMPT}

Generate step-by-step crochet instructions for this image.

[Image]
\end{tcolorbox}

\subsection{Task D (Step-level): NL $\rightarrow$ DSL Translation Prompt}
This task evaluates whether a model can translate a single natural language instruction 
into exactly one line of compilable \textbf{CrochetPARADE} DSL code.

\begin{tcolorbox}[title=Step-level NL $\rightarrow$ DSL Translation Prompt (Rendered Example),
                  breakable, colback=white, colframe=black, sharp corners=south,
                  before skip=8pt, after skip=8pt,
                  listing only,
                  listing options={basicstyle=\ttfamily\footnotesize,breaklines=true,columns=fullflexible}]
\textbf{SYSTEM PROMPT}

You are a crochet compiler. Translate the next instruction NL into one line of CrochetPARADE DSL.

Use consistent naming and syntax.

Important rules for translations:

1. Make sure your output ONLY contains the DSL code, nothing else.

2. Use the previous examples to understand the pattern of translation.

3. Be consistent in naming conventions with the examples.

4. Your output should be exactly one line of DSL code.

\vspace{1em} 

\textbf{USER PROMPT}

Now translate the NL into DSL:

NL:

DSL:
\end{tcolorbox}

\subsection{Task D (Project-Level): NL $\rightarrow$ DSL Translation Prompt}
This task evaluates whether a model can convert natural language crochet instructions (with optional images) into compilable CrochetPARADE DSL code.

\begin{tcolorbox}[title=NL $\rightarrow$ DSL Translation Prompt (Rendered Example), breakable,  colback=white, colframe=black, sharp corners=south]{
\textbf{SYSTEM PROMPT}

You are a professional crochet pattern writer.
Convert instructions + images into compilable CrochetPARADE DSL code.
Output only the DSL code. No explanations, commentary, or extra text.

Example 1:

\quad  "image path": \url{https://www.yarnspirations.com/cdn/shop/files/BRC0116-035467M.jpg},

\vspace{1em} 
  
INSTRUCTIONS
 
\quad Note:  Join with sl st to first sc at end of each rnd.

\quad Ch 2.

\quad**Rnd 1:** 6 sc in 2nd ch from hook. Join. (6 sc)

\quad**Rnd 2:** Ch 1. 2 sc in each sc around. Join. (12 sc)

\quad**Rnd 3:** Ch 1. (2 sc in next sc, 1 sc in next sc) repeat around. End with 1 sc. Join. (18 sc)

\quad**Rnd 4:** Ch 1. (2 sc in next sc, 1 sc in each of next 2 sc) repeat. End with 1 sc in last 2 sc. Join. (24 sc)

\quad**Rnd 5:** Ch 1. Sc in each sc around. Join. (24 sc)

\quad**Rnd 6:** Ch 1. (2 sc in next sc, 1 sc in each of next 3 sc) repeat. End with 1 sc in last 3 sc. Join. (30 sc)

\quad**Rnds 7–8:** Repeat Rnd 5 (sc in each sc). Join. (30 sc each round)

\quad**Rnd 9:** Ch 1. **Working in back loops only**: (2 sc in next sc, 1 sc in each of next 2 sc) repeat. End with 1 sc in last 2 sc. Join. (40 sc)

\quad**Rnd 10:** Ch 1. Sc in each sc around (both loops). Join. (40 sc)

\quad**Rnd 11:** Ch 1. (2 sc in next sc, 1 sc in each of next 3 sc) repeat. End with 1 sc in last 3 sc. Join. (50 sc)

\quad**Finish:** Fasten off.

\vspace{1em} 

DSL

¶ch.B

¶\$k=0\$,sc@B.A[k],5sc@B,ss@A[k++]

¶ch.A[k],sk,6sc2inc,ss@A[k++]

¶ch.A[k],sk,[sc2inc,sc]*6,ss@A[k++]

¶ch.A[k],sk,[sc2inc,2sc]*6,ss@A[k++]

¶ch.A[k],sk,24sc,ss@A[k++]

¶ch.A[k],sk,[sc2inc,3sc]*6,ss@A[k++]

¶[ch.A[k],sk,30sc,ss@A[k++]

¶]*2

¶ch.A[k],sk,[scbl,scbl@[@],2scbl]*10,
ss@A[k++]

¶ch.A[k],sk,40sc,ss@A[k++]

¶ch.A[k],sk,[sc2inc,3sc]*10,ss@A[k++]

¶ch.A[k],sk,[2sc,3sc]*10,ss@A[k++]

\vspace{1em} 

\textbf{USER PROMPT}

Now generate DSL code for the following:

\vspace{1em} 

[Image]  
\vspace{1em}

\vspace{1em} 

[Instructions]  

Rnd 1: Ch 2, 6 sc in ring  

Rnd 2: 2 sc in each (12)  

Rnd 3: [Sc, sc, inc] around (16)  

Rnd 4: [Tr, sc] repeat around

\vspace{1em} 

[DSL]
}
\end{tcolorbox}

%% file: appendix/bertscore.tex
\section{BERTScore Analysis}
\label{app:bert_score}

We compute BERTScore (Precision, Recall, F1) for all Task~C outputs. As shown in Table~\ref{tab:bertscore}, BERTScore F1 values cluster within a relatively narrow range (0.80–0.83 for most strong models), even when BLEU/ROUGE scores are low and many outputs fail under executable evaluation in Task~D. This suggests that embedding-based similarity captures surface-level fluency and semantic plausibility, but provides limited discrimination between procedurally correct and structurally invalid instructions.

We attribute this behavior to the nature of crochet instructions. The language is highly templated and lexically repetitive, with a small vocabulary of stitch operations and recurring syntactic patterns. As a result, embedding similarity primarily reflects topical coherence (i.e., whether an output “looks like” a crochet pattern), while remaining relatively insensitive to structurally critical token-level differences, such as stitch counts, increase/decrease operations, repeat multipliers, and round-level dependencies. For instance, “3 dc in next st” and “2 dc in next st” are semantically similar under contextual embeddings, yet lead to fundamentally different procedural outcomes.

\begin{table}[H]
\centering
\small
\begin{tabular}{lccc}
\toprule
Model & BS-P & BS-R & BS-F1 \\
\midrule
BLIP & 0.8219 & 0.7811 & 0.8009 \\
Gemma3-4B & 0.5243 & 0.4939 & 0.5079 \\
Claude & 0.8306 & 0.8313 & 0.8309 \\
DeepSeek-VL & 0.8035 & 0.7944 & 0.7987 \\
Qwen2-VL-7B & 0.8053 & 0.7954 & 0.8002 \\
Gemini & 0.8111 & 0.8327 & 0.8217 \\
GPT-4o & 0.8181 & 0.8271 & 0.8224 \\
\bottomrule
\end{tabular}
\caption{BERTScore results on Task~C. BS denotes BERTScore.}
\label{tab:bertscore}
\end{table}
In contrast, character- and token-level metrics (BLEU, ROUGE, ChrF), while imperfect, are more sensitive to stitch-level variations and count discrepancies. Nevertheless, textual similarity alone remains insufficient to capture procedural correctness. CrochetBench therefore emphasizes Task~D’s executor-based evaluation, where DSL compilability and rendering-based DINO similarity assess whether generated instructions produce valid and structurally faithful outcomes. Since crochet procedures are not unique and multiple valid derivations may exist, outcome-based validation provides a more reliable signal than reference-aligned textual similarity.

%% file: appendix/crochetparade.tex
\section{CrochetPARADE: Pattern Renderer, Analyzer, and Debugger}
\label{appendix:crochetparade}
\begin{figure*}[ht]
\centering
\includegraphics[width=0.23\textwidth]{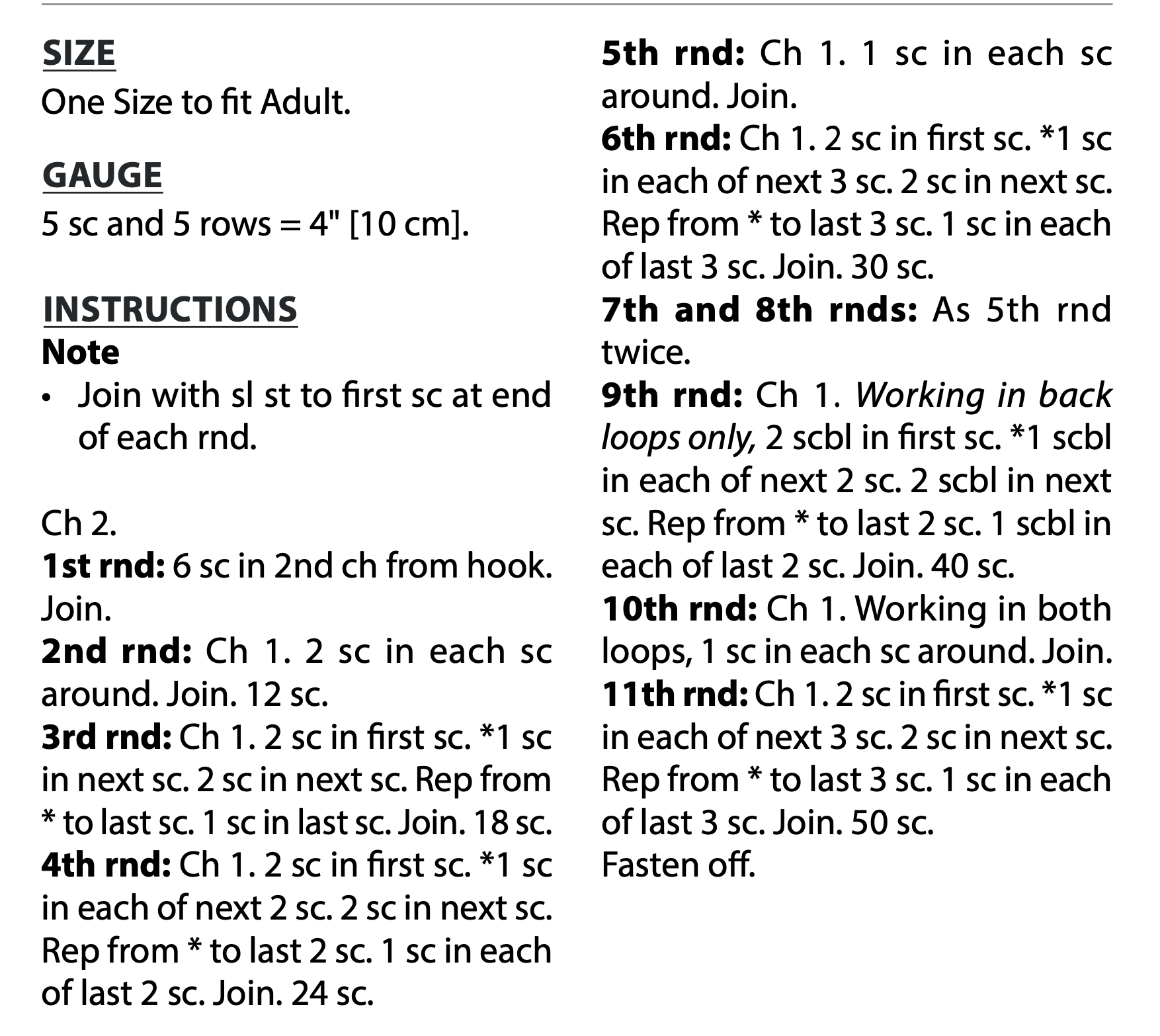} 
\includegraphics[width=0.23\textwidth]{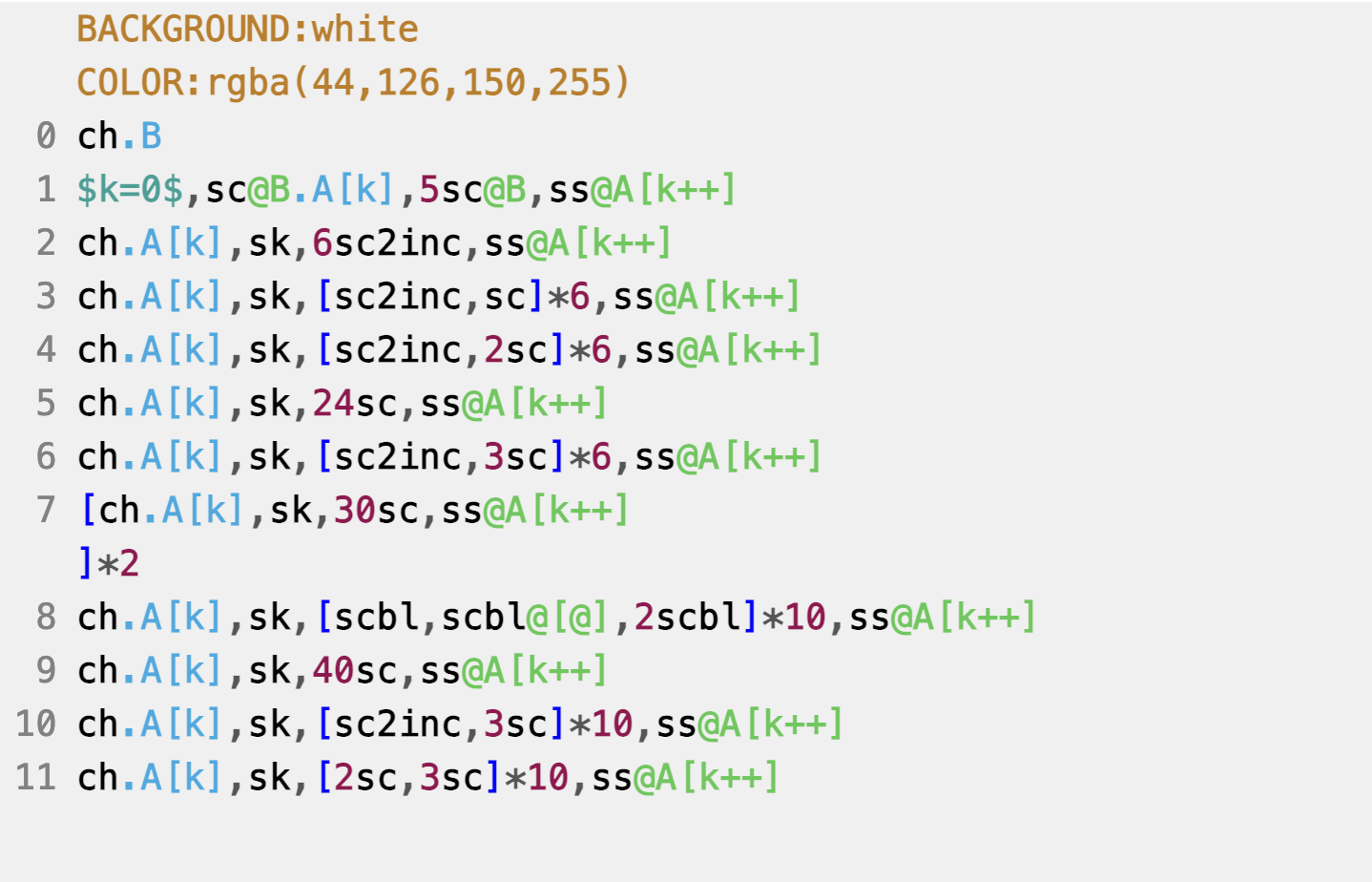}     
\includegraphics[width=0.23\textwidth]{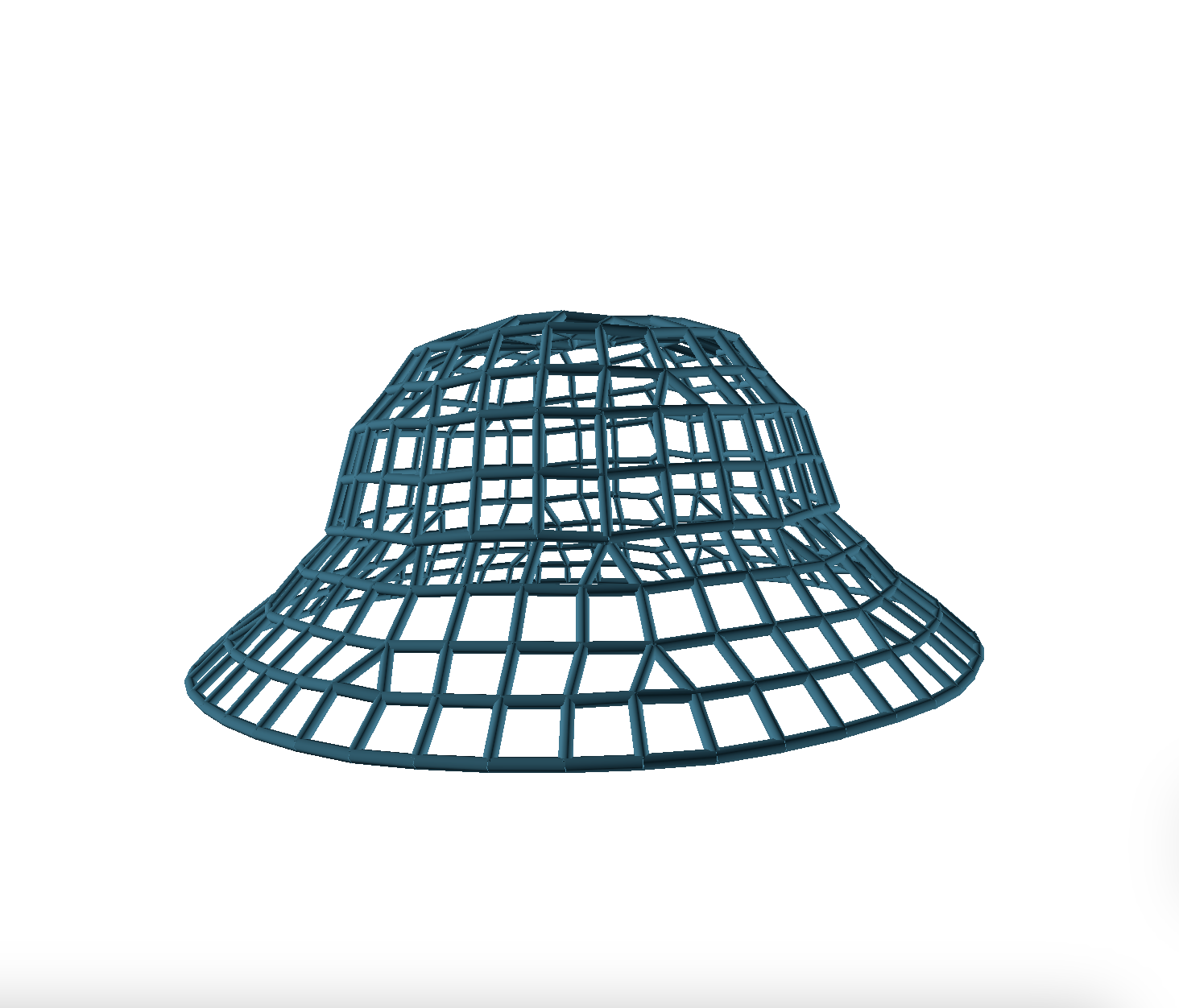}   
\includegraphics[width=0.23\textwidth]{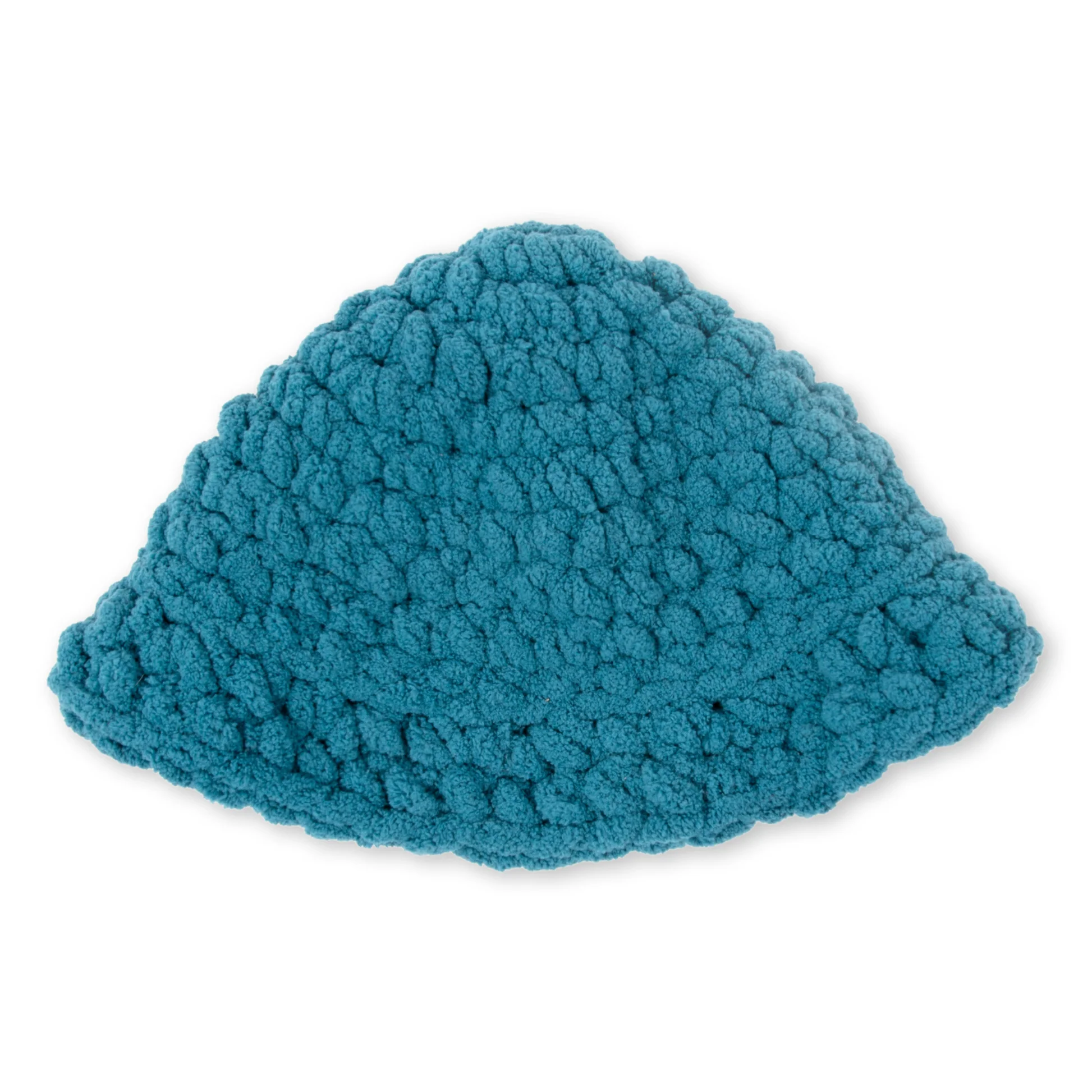}   
\caption{Example of the CrochetBench translation pipeline. 
(Left) Natural language crochet instructions from the dataset. 
(Second) Automatically translated into CrochetPARADE DSL, a formal stitch grammar. 
(Third) Mesh rendering generated from the DSL. 
(Right) Target crocheted item image provided in the dataset. 
This pipeline enables direct text-to-image consistency checks, automated validation, 
and future training of NL $\rightarrow$ DSL models, analogous to text-to-code generation.}
\label{fig:crochetbench-pipeline}
\end{figure*}

\textbf{CrochetPARADE} (short for \emph{Crochet Pattern Renderer, Analyzer, and Debugger}) is an interactive platform that enables users to author, visualize, test, and export crochet patterns in both 2D and 3D \citep{crochetparade}. By combining a custom pattern grammar with simulation and rendering tools, CrochetPARADE addresses common issues of ambiguity, correctness, and interpretability in textual crochet instructions.\footnote{\url{https://www.crochetparade.org/}}

\paragraph{Core Capabilities.}  
\begin{itemize}
    \item \textbf{Interactive authoring and rendering.} Users write pattern instructions in the CrochetPARADE grammar and then invoke a ``calculate'' operation to convert those instructions into a virtual model. The system supports both 2D and 3D views, along with interactive controls such as zoom, rotation, and stitch highlighting.
    \item \textbf{Validation and debugging.} CrochetPARADE parses the input, checks for syntactic and consistency errors (e.g., mismatched stitch counts, impossible attachments), and flags over- or under-stretched stitches.
    \item \textbf{Export and interoperability.} From a rendered pattern, users can export:
    \begin{itemize}
        \item A standard crochet chart (SVG) with conventional stitch symbols and labeled stitch connections.  
        \item A 3D model (GLTF format) for integration into external tools such as Blender.  
        \item The underlying pattern instructions text (in the CrochetPARADE grammar), ensuring reproducibility and sharing.  
    \end{itemize}
\end{itemize}

\paragraph{Design Ideals and Rationale.}  
CrochetPARADE is built to meet several design goals: (i) \emph{unambiguous precision}, where the grammar is far more strict than free-form natural language, reducing interpretive errors; (ii) \emph{local computation}, since all parsing, simulation, and rendering occur client-side in the browser with no user instructions sent to a central server; and (iii) \emph{open source extensibility}, as the platform is released under GPLv3, with the grammar manual provided under a Creative Commons BY-NC-SA license.

\paragraph{Role in Our Work.}  
Within the context of CrochetBench, CrochetPARADE provides a rigorous target representation: model predictions can be compiled into CrochetPARADE instructions, validated for syntactic and structural correctness, and then visualized or executed. This enables evaluation beyond surface-level metrics (e.g., BLEU, ROUGE) toward \emph{executor correctness}---whether a generated pattern is valid, renderable, and stitch-balanced.

%% file: appendix/error_type.tex
\section{DSL Error Taxonomy}
\label{app:dsl_error}
To better understand failure cases in Task D, we extend the validator’s error analysis 
with detailed subcategories and examples. 

\begin{description}
  \item[Unbalanced Brackets.] Missing opening/closing parentheses or brackets.  
  \begin{tcolorbox}[colback=white, colframe=black!50, sharp corners=south, breakable, title=Examples]
  \texttt{Unbalanced brackets: (sc,hc5,sltr)infl)} \\
  \end{tcolorbox}

  \item[Multiple References.] Improper formatting of references.  
  \begin{tcolorbox}[colback=white, colframe=black!50, sharp corners=south, breakable, title=Example]
  \texttt{Multiple references defined without parenthesis: (21ch),turn\\
  sk,(20sc)\\
  (2ndrow):Ch1.(1scbl)
  ineachchtoendofrow.Turn}
  \end{tcolorbox}

  \item[Undefined Stitch.]
Undefined stitch types not in the dictionary.
\begin{tcolorbox}[colback=white, colframe=black!50, sharp corners=south, breakable, title=Examples]
\texttt{ch1}, \texttt{ch3}, \texttt{scfp}, \texttt{hdc\_bar}
\end{tcolorbox}
  \item[Variable Naming Conflict.] Conflicts between variable names and stitch names.  
  \begin{tcolorbox}[colback=white, colframe=black!50, sharp corners=south, title=Example]
  \texttt{Error: variable name matches stitch name. For example, \$ch=0\$ cannot be used since 'ch' is a stitch name.}
  \end{tcolorbox}

  \item[Label Not Found.] Reference to a non-existent label.  
  \begin{tcolorbox}[colback=white, colframe=black!50, sharp corners=south, title=Example]
  \texttt{Label not found: C}
  \end{tcolorbox}

  \item[Non-Adjacent Labels.] Same label used for non-adjacent stitches.  
  \begin{tcolorbox}[colback=white, colframe=black!50, sharp corners=south, title=Example]
  \texttt{Cannot use same label over non-adjacent stitches. Consider using different labels.}
  \end{tcolorbox}

  \item[Turning Issue.] Misplaced turning commands.  
  \begin{tcolorbox}[colback=white, colframe=black!50, sharp corners=south, title=Example]
  \texttt{Turning can happen only at the end of a row.}
  \end{tcolorbox}

  \item[Runtime Errors.] Low-level parsing failures from the JavaScript compiler.  
  \begin{tcolorbox}[colback=white, colframe=black!50, sharp corners=south, breakable, title=Examples]
  \texttt{Cannot read properties of null (reading '0')} \\
  \texttt{Cannot use 'in' operator to search for 'attach\_id' in NaN}
  \end{tcolorbox}

  \item[Multiplier Issue.] Improper formatting in multiplier.
\begin{tcolorbox}[colback=white, colframe=black!50, sharp corners=south, breakable, title=Examples]
\texttt{Error: Exception during pattern parsing: Multiplier set, but no stitch found: ch.B}
\end{tcolorbox}
\end{description}

%% file: appendix/case_study_example.tex
\section{Example Model-Generated Crochet Instructions}
\label{app:claude_example}

This appendix provides the full verbatim crochet instructions generated by each model for the Task~C (Instruction Generation) case study (Figure~\ref{fig:case_study_main}). We include the complete outputs without truncation or post-editing to preserve long-horizon procedural dependencies and enable transparent qualitative inspection. Because many failure modes in crochet pattern generation only emerge after multiple rounds, partial excerpts can obscure structural inconsistencies and topology errors. Presenting the full instructions allows readers to directly assess how each model handles stateful pattern progression, global geometry, and motif repetition beyond surface-level textual fluency.

\begin{tcolorbox}[
  breakable,
  enhanced,
  colback=white,
  colframe=black,
  boxrule=0.6pt,
  arc=3pt,
  left=6pt,
  right=6pt,
  top=6pt,
  bottom=6pt,
  title=\textbf{Ground Truth},
  fonttitle=\bfseries,
]

\footnotesize
\medskip
\textbf{Notes:}
\begin{itemize}
    \item Ch 2 at beg of rnd does not count as hdc.
    \item Ch 3 at beg of rnd counts as dc.
    \item To change colors, work to last 2 loops on hook of first color. Draw new color through last 2 loops and proceed with new color.
\end{itemize}
 1. Ch 2 at beg of rnd does not count as hdc.



\medskip
\textbf{Stripe Pat:} \\
With A, work 4 rnds. \\
With B, work 4 rnds. \\
These 8 rnds form Stripe Pat.

\medskip
Ch 4. Join with sl st to first ch to form ring. See diagram on page 3.

\medskip
\textbf{1st rnd:} Ch 3. 13 dc in ring. Join with sl st to top of ch 3. 14 dc.

\textbf{2nd rnd:} Ch 3. 1 dc in same sp as last sl st. 2 dc in each dc around. Join with sl st to top of ch 3. 28 dc.

\textbf{3rd rnd:} Ch 3. 4 dc in same sp as last sl st. \\
*Dc3tog. 5 dc in next dc.* \\
Rep from * to last 3 sts. Dc3tog. Join with sl st to top of ch 3.

\textbf{4th rnd:} Sl st in next dc. Ch 2. 3 hdc in same dc as sl st. \\
*Bobble in next st. 3 hdc in next dc. Hdc3tog.* \\
3 hdc in next dc. Rep from * around, ending last rep at **. Join with sl st to first hdc.

\textbf{5th rnd:} Sl st in next st. Ch 3. \\
*3 dc in next st. 1 dc in next st. 3 dc in next st. 1 dc in next st. Dc3tog.* \\
1 dc in next st. Rep from * around, ending last rep at **. Join with sl st to top of ch 3.

\textbf{6th rnd:} Sl st in next st. Ch 2. 1 hdc in same st as sl st. \\
*(Bobble in next st. 3 hdc in next st) twice. Bobble in next st. 1 hdc in next st. Hdc3tog.* \\
1 hdc in next st. Rep from * around, ending last rep at **. Join with sl st to first hdc.

\textbf{7th rnd:} Sl st in next st. Ch 3. 1 dc in each of next 2 sts. \\
*3 dc in next st. 1 dc in next st. 3 dc in next st. 1 dc in each of next 3 sts. Dc3tog.* \\
1 dc in each of next 3 sts. Rep from * around, ending last rep at **. Join with sl st to top of ch 3.

\textbf{8th rnd:} Sl st in next st. Ch 2. 1 hdc in same st as sl st. \\
*Bobble in next st. 1 hdc in next st. (Bobble in next st. 3 hdc in next st) twice. (Bobble in next st. 1 hdc in next st) twice. Hdc3tog.* \\
1 hdc in next st. Rep from * around, ending last rep at **.

\textbf{9th rnd:} Sl st in next st. Ch 3. 1 dc in each of next 4 sts. \\
3 dc in next st. 1 dc in next st. PM in dc. 3 dc in next st. \\
1 dc in each of next 5 sts. Dc3tog. PM in dc3tog. \\
*1 dc in each of next 5 sts. 3 dc in next st. 1 dc in next st. PM in dc. \\
3 dc in next st. 1 dc in each of next 5 sts. Dc3tog. PM in dc3tog.* \\
Rep from * around. Join with sl st to top of ch 3.

\textbf{10th rnd:} Sl st in next st. \\
*1 hdc in next st. Bobble in next st.* Rep from * to * until 1 st before next marked st. \\
3 hdc in next st. Bobble in next st. PM in Bobble. 3 hdc in next st. \\
*Bobble in next st. 1 hdc in next st.* Rep from ** to 1 st before next marked st. \\
Hdc3tog. PM in hdc3tog. Rep from * around. Join with sl st to first hdc.

\textbf{11th rnd:} Sl st in next st. Ch 3. \\
*1 dc in each st to 1 st before marked st. 3 dc in next st. 1 dc in next st. PM in dc. \\
3 dc in next st. 1 dc in each st until 1 st before next marked st. Dc3tog. PM in dc3tog.* \\
Rep from * around. Join with sl st to top of ch 3.

\medskip
Rep 10th and 11th rnds, keeping cont of Stripe Pat, until work from center to outer point measures approx 30" [76 cm]. Fasten off.

\medskip
\textbf{Tassels (make 7):} \\
Cut a piece of cardboard 5" [12.5 cm] wide. Wind B around cardboard 40 times. Break yarn leaving a long end and thread end through needle. Slip needle through all loops and tie tightly. Remove cardboard and wind yarn tightly around loops 1" [2.5 cm] below fold. Fasten securely. Cut through remaining loops and trim ends evenly. Sew one tassel to each ``peak'' along the edge of Blanket as pictured.

\end{tcolorbox}


\begin{tcolorbox}[
  breakable,
  enhanced,
  colback=white,
  colframe=black,
  boxrule=0.6pt,
  arc=3pt,
  left=6pt,
  right=6pt,
  top=6pt,
  bottom=6pt,
  title=\textbf{Gemini},
  fonttitle=\bfseries,
]

\footnotesize

\textbf{Special Stitch}

\textbf{Bobble:} [Yarn over, insert hook in indicated st, yarn over and pull up a loop, yarn over and draw through 2 loops] 5 times in same st, yarn over and draw through all 6 loops on hook.

\medskip
\textbf{Instructions}

With Color A (Blue), make a magic ring.

\textbf{Rnd 1:} Ch 1, 14 sc in ring. Join with sl st to first sc. [14 sc]

\textbf{Rnd 2:} Ch 1, sc in first st. *Ch 2, skip 1 st, sc in next st. Rep from * around to last st, ch 2, skip last st. Join with sl st to first sc. Fasten off A. [7 sc, 7 ch-2 sps]

\medskip
Join Color B (Brown) with sl st in any ch-2 sp.

\textbf{Rnd 3:} Ch 1, (Bobble, ch 2, Bobble) in same sp. *Sc in next sc, (Bobble, ch 2, Bobble) in next ch-2 sp. Rep from * around. Join with sl st to top of first Bobble. [14 Bobbles, 7 sc]

\textbf{Rnd 4:} Sl st into ch-2 sp. Ch 1, (Bobble, ch 2, Bobble) in same sp. *Sc in next Bobble, Bobble in next sc, sc in next Bobble. (Bobble, ch 2, Bobble) in next ch-2 sp. Rep from * around, omitting last point group on final rep. Join with sl st to top of first Bobble. Fasten off B. [3 sts on each of 7 sides]

\medskip
Join Color A with sl st in any ch-2 sp.

\textbf{Rnd 5:} Ch 1, (Bobble, ch 2, Bobble) in same sp. *(Sc in next st, Bobble in next st) twice, sc in next st. (Bobble, ch 2, Bobble) in next ch-2 sp. Rep from * around, omitting last point group on final rep. Join with sl st to top of first Bobble. [5 sts on each of 7 sides]

\textbf{Rnd 6:} Sl st into ch-2 sp. Ch 1, (Bobble, ch 2, Bobble) in same sp. *(Sc in next st, Bobble in next st) three times, sc in next st. (Bobble, ch 2, Bobble) in next ch-2 sp. Rep from * around, omitting last point group on final rep. Join with sl st to top of first Bobble. Fasten off A. [7 sts on each of 7 sides]

\medskip
Join Color B with sl st in any ch-2 sp.

\textbf{Rnd 7:} Ch 1, (Bobble, ch 2, Bobble) in same sp. *(Sc in next st, Bobble in next st) four times, sc in next st. (Bobble, ch 2, Bobble) in next ch-2 sp. Rep from * around, omitting last point group on final rep. Join with sl st to top of first Bobble. [9 sts on each of 7 sides]

\textbf{Rnd 8:} Sl st into ch-2 sp. Ch 1, (Bobble, ch 2, Bobble) in same sp. *(Sc in next st, Bobble in next st) five times, sc in next st. (Bobble, ch 2, Bobble) in next ch-2 sp. Rep from * around, omitting last point group on final rep. Join with sl st to top of first Bobble. Fasten off B. [11 sts on each of 7 sides]

\medskip
Join Color A with sl st in any ch-2 sp.

\textbf{Rnd 9:} Ch 1, (Bobble, ch 2, Bobble) in same sp. *(Sc in next st, Bobble in next st) six times, sc in next st. (Bobble, ch 2, Bobble) in next ch-2 sp. Rep from * around, omitting last point group on final rep. Join with sl st to top of first Bobble. [13 sts on each of 7 sides]

\textbf{Rnd 10:} Sl st into ch-2 sp. Ch 1, (Bobble, ch 2, Bobble) in same sp. *(Sc in next st, Bobble in next st) seven times, sc in next st. (Bobble, ch 2, Bobble) in next ch-2 sp. Rep from * around, omitting last point group on final rep. Join with sl st to top of first Bobble. Fasten off. [15 sts on each of 7 sides]

\medskip
\textbf{Finishing}

Weave in ends. \\
Make 7 tassels with Color B and attach one to each ch-2 point of the star.

\end{tcolorbox}

\begin{tcolorbox}[
  breakable,
  enhanced,
  colback=white,
  colframe=black,
  boxrule=0.6pt,
  arc=3pt,
  left=6pt,
  right=6pt,
  top=6pt,
  bottom=6pt,
  title=\textbf{GPT-4o},
  fonttitle=\bfseries,
]

\footnotesize

\textbf{Star-Shaped Crochet Blanket Pattern}

\medskip
\textbf{Materials:}
\begin{itemize}
  \item Yarn: Brown and Blue
  \item Crochet hook: Size H/8 (5.00 mm)
  \item Yarn needle
  \item Scissors
  \item Optional: Tassels (for points)
\end{itemize}

\medskip
\textbf{Abbreviations:}
\begin{itemize}
  \item ch = chain
  \item sc = single crochet
  \item hdc = half double crochet
  \item dc = double crochet
  \item tr = treble
  \item sl st = slip stitch
  \item rep = repeat
\end{itemize}

\medskip
\textbf{Instructions}

\medskip
\textbf{Foundation Ring:} \\
With Brown yarn, make a magic ring.

\medskip
\textbf{Rnd 1:} \\
Ch 3 (counts as dc), 11 dc in ring. Join with sl st to top of ch-3. (12 dc)

\textbf{Rnd 2:} \\
Ch 3, dc in same space, 2 dc in each st around. Join. (24 dc)

\textbf{Rnd 3:} \\
Ch 3, *dc in next st, 2 dc in next st; rep from * around. Join. (36 dc)

\textbf{Rnd 4:} \\
Ch 3, *dc in next 2 sts, 2 dc in next st; rep from * around. Join. (48 dc)

\medskip
\textbf{Change to Blue Yarn.}

\medskip
\textbf{Rnd 5:} \\
Ch 3, *dc in next 3 sts, 2 dc in next st; rep from * around. Join. (60 dc)

\textbf{Rnd 6:} \\
Ch 3, *dc in next 4 sts, 2 dc in next st; rep from * around. Join. (72 dc)

\textbf{Rnd 7:} \\
Ch 3, *dc in next 5 sts, 2 dc in next st; rep from * around. Join. (84 dc)

\medskip
\textbf{Change to Brown Yarn.}

\medskip
\textbf{Rnd 8:} \\
Ch 3, *dc in next 6 sts, 2 dc in next st; rep from * around. Join. (96 dc)

\textbf{Rnd 9--11:} \\
Repeat Rnds 5--7 (using Blue \& Brown alternately).

\medskip
\textbf{Finishing:} \\
Fasten off and weave in ends.

\medskip
\textbf{Tassels:}
\begin{itemize}
  \item Cut 4 lengths of Brown yarn (about 8 inches each).
  \item Wrap yarn around fingers, tie off center, trim ends, and attach one tassel to each point of the star.
\end{itemize}

\medskip
\textbf{Block if necessary.}

\medskip
\textbf{Enjoy your star-shaped blanket!}

\end{tcolorbox}

\begin{tcolorbox}[
  breakable,
  enhanced,
  colback=white,
  colframe=black,
  boxrule=0.6pt,
  arc=3pt,
  left=6pt,
  right=6pt,
  top=6pt,
  bottom=6pt,
  title=\textbf{Claude},
  fonttitle=\bfseries,
]

\small
\textbf{8-Point Star Afghan}

\textbf{Materials:} Worsted weight yarn in blue and brown \\
\textbf{Hook:} Size H/8 (5mm)

\textbf{Center:} \\
With blue, ch 4, join with sl st to form ring.

\textbf{Rnd 1:} Ch 1, 8 sc in ring, join with sl st. (8 sc) \\
\textbf{Rnd 2:} Ch 1, 2 sc in each sc around, join. (16 sc) \\
\textbf{Rnd 3:} Ch 1, *sc in next sc, 2 sc in next sc*, rep around, join. (24 sc) \\
\textbf{Rnd 4:} Ch 1, *sc in next 2 sc, 2 sc in next sc*, rep around, join. (32 sc) \\
\textbf{Rnd 5:} Ch 1, *sc in next 3 sc, 2 sc in next sc*, rep around, join. (40 sc) \\
\textbf{Rnd 6:} Ch 1, *sc in next 4 sc, 2 sc in next sc*, rep around, join. (48 sc)

\medskip
\textbf{Star Points Formation:}

\textbf{Rnd 7:} Ch 1, *sc in next 5 sc, 2 sc in next sc*, rep around, join. (56 sc) \\
\textbf{Rnd 8:} Ch 1, *sc in next 6 sc, 2 sc in next sc*, rep around, join. (64 sc) \\
\textbf{Rnd 9:} Change to brown. Ch 1, sc in each sc around, join. \\
\textbf{Rnd 10:} Ch 1, *sc in next 7 sc, 2 sc in next sc*, rep around, join. (72 sc) \\
\textbf{Rnd 11:} Ch 1, *sc in next 8 sc, 2 sc in next sc*, rep around, join. (80 sc) \\
\textbf{Rnd 12:} Ch 1, *sc in next 9 sc, 2 sc in next sc*, rep around, join. (88 sc) \\
\textbf{Rnd 13:} Change to blue. Ch 1, sc in each sc around, join. \\
\textbf{Rnd 14:} Ch 1, *sc in next 10 sc, 2 sc in next sc*, rep around, join. (96 sc) \\
\textbf{Rnd 15:} Ch 1, *sc in next 11 sc, 2 sc in next sc*, rep around, join. (104 sc) \\
\textbf{Rnd 16:} Ch 1, *sc in next 12 sc, 2 sc in next sc*, rep around, join. (112 sc) \\
\textbf{Rnd 17:} Change to brown. Ch 1, sc in each sc around, join. \\
\textbf{Rnd 18:} Ch 1, *sc in next 13 sc, 2 sc in next sc*, rep around, join. (120 sc) \\
\textbf{Rnd 19:} Ch 1, *sc in next 14 sc, 2 sc in next sc*, rep around, join. (128 sc) \\
\textbf{Rnd 20:} Change to blue. Ch 1, sc in each sc around, join.

\medskip
Continue pattern of color changes every 3--4 rounds, increasing evenly to maintain flat star shape until desired size.

\medskip
\textbf{Edging:} \\
Final Rnd: With blue, ch 1, *sc in next 2 sc, ch 3, sl st in 3rd ch from hook (picot made)*, rep around, join.

\medskip
\textbf{Tassels:} \\
Cut 40 strands of brown yarn, 4 inches long. Divide into 8 groups of 5 strands each. Attach one tassel to each star point.

\end{tcolorbox}

\begin{tcolorbox}[
  breakable,
  enhanced,
  colback=white,
  colframe=black,
  boxrule=0.6pt,
  arc=3pt,
  left=6pt,
  right=6pt,
  top=6pt,
  bottom=6pt,
  title=\textbf{Qwen-VL-7B},
  fonttitle=\bfseries,
]

\footnotesize

\textbf{Star Motif Crochet Pattern}

\medskip
\textbf{Materials:}
\begin{itemize}
  \item Worsted weight yarn in navy blue and brown
  \item Crochet hook size H (5.0 mm) or size needed to obtain gauge
  \item Yarn needle for weaving in ends
  \item Tassels (optional)
\end{itemize}

\medskip
\textbf{Gauge:} 
16 sc and 16 hdc around a 4" circle

\medskip
\textbf{Instructions:}

\medskip
\textbf{Round 1:} 
Ch 10, join with sl st to form a ring. \\
10 sc in the ring.

\medskip
\textbf{Round 2:} 
Ch 1, 1 sc in each sc around. \\
Fasten off.

\medskip
\textbf{Round 3:} 
Ch 1, 1 hdc in each sc around. \\
Fasten off.

\medskip
\textbf{Round 4:} 
Ch 1, 1 dc in each hdc around. \\
Fasten off.

\medskip
\textbf{Round 5:} 
Ch 1, 1 tr in each dc around. \\
Fasten off.

\medskip
\textbf{Round 6:} 
Ch 1, 1 hdc in each tr around. \\
Fasten off.

\medskip
\textbf{Round 7:} 
Ch 1, 1 sc in each hdc around. \\
Fasten off.

\medskip
\textbf{Round 8:} 
Ch 1, 1 hdc in each sc around. \\
Fasten off.

\medskip
\textbf{Round 9:} 
Ch 1, 1 dc in each hdc around. \\
Fasten off.

\medskip
\textbf{Round 10:} 
Ch 1, 1 tr in each dc around. \\
Fasten off.

\medskip
\textbf{Round 11:} 
Ch 1, 1 hdc in each tr around. \\
Fasten off.

\medskip
\textbf{Round 12:} 
Ch 1, 1 sc in each hdc around. \\
Fasten off.

\medskip
\textbf{Finishing:} \\
Join the rounds with slip stitches to form a star shape. \\
Weave in ends. \\
Attach tassels to each point of the star (optional).

\medskip
\textbf{Note:} Adjust the number of rounds to achieve the desired size of the star.

\end{tcolorbox}

\begin{tcolorbox}[
  breakable,
  enhanced,
  colback=white,
  colframe=black,
  boxrule=0.6pt,
  arc=3pt,
  left=6pt,
  right=6pt,
  top=6pt,
  bottom=6pt,
  title=\textbf{DeepSeek-VL},
  fonttitle=\bfseries,
]

\footnotesize

ch 100, sc in 2nd ch from hook, sc in each ch across, join with sl st in 1st sc, 100 loops made.

\end{tcolorbox}

\begin{tcolorbox}[
  breakable,
  enhanced,
  colback=white,
  colframe=black,
  boxrule=0.6pt,
  arc=3pt,
  left=6pt,
  right=6pt,
  top=6pt,
  bottom=6pt,
  title=\textbf{Qwen-VL-72B},
  fonttitle=\bfseries,
]

\footnotesize

\textbf{Star Blanket Crochet Pattern}

\medskip
\textbf{Materials:}
\begin{itemize}
  \item Yarn in two colors: Blue (A) and Brown (B)
  \item Crochet hook size H/8 (5mm)
\end{itemize}

\medskip
\textbf{Abbreviations:}
\begin{itemize}
  \item ch = chain
  \item sc = single crochet
  \item hdc = half double crochet
  \item dc = double crochet
  \item sl st = slip stitch
\end{itemize}

\medskip
\textbf{Pattern}

\medskip
\textbf{Foundation Round:} \\
With Blue (A), ch 4, join with sl st to form a ring.

\medskip
\textbf{Round 1:} \\
Ch 3 (counts as first dc), work 2 dc in ring, *ch 1, skip next ch, work 3 dc in ring; rep from * around, join with sl st to top of beg ch-3. (6 clusters)

\medskip
\textbf{Round 2:} \\
Ch 3 (counts as first dc), work 2 dc in same st, *ch 1, skip next ch, work 3 dc in next cluster; rep from * around, join with sl st to top of beg ch-3. (12 clusters)

\medskip
\textbf{Round 3:} \\
Ch 3 (counts as first dc), work 2 dc in same st, *ch 1, skip next ch, work 3 dc in next cluster; rep from * around, join with sl st to top of beg ch-3. (24 clusters)

\medskip
\textbf{Round 4:} \\
Ch 3 (counts as first dc), work 2 dc in same st, *ch 1, skip next ch, work 3 dc in next cluster; rep from * around, join with sl st to top of beg ch-3. (48 clusters)

\medskip
\textbf{Round 5:} \\
Ch 3 (counts as first dc), work 2 dc in same st, *ch 1, skip next ch, work 3 dc in next cluster; rep from * around, join with sl st to top of beg ch-3. (96 clusters)

\medskip
\textbf{Color Change:} \\
Join Brown (B) with sl st in next ch-1 sp.

\medskip
\textbf{Round 6:} \\
Ch 3 (counts as first dc), work 2 dc in same st, *ch 1, skip next ch, work 3 dc in next cluster; rep from * around, join with sl st to top of beg ch-3. (192 clusters)

\medskip
\textbf{Round 7:} \\
Ch 3 (counts as first dc), work 2

\medskip

\end{tcolorbox}